\documentclass[sn-mathphys,Numbered]{sn-jnl}% Math and Physical Sciences Reference Style
\usepackage{graphicx}%
\usepackage{multirow}%
\usepackage{amsmath,amssymb,amsfonts}%
\usepackage{amsthm}%
\usepackage{mathrsfs}%
\usepackage[title]{appendix}%
\usepackage{xcolor}%
\usepackage{textcomp}%
\usepackage{manyfoot}%
\usepackage{booktabs}%
\usepackage{algorithm}%
\usepackage{algorithmicx}%
\usepackage{algpseudocode}%
\usepackage{listings}%
\usepackage[utf8]{inputenc}
\usepackage{hyperref}
\usepackage{pdfpages}

\begin{document}
\title[Direct mineral content prediction from drill core images via transfer learning]{Direct mineral content prediction from drill core images via transfer learning}
%\title[Transfer learning based mineral content #\\
%regression from drill core images]{Transfer 
%learning based mineral content \\
%regression from drill core images}


\author*[1]{\fnm{Romana} \sur{Boiger}}\email{romana.boiger@psi.ch}

\author[1,2]{\fnm{Sergey V.} \sur{Churakov}}\email{sergey.churakov@psi.ch}

\author[1,3]{\fnm{Ignacio} \sur{Ballester Llagaria}}\email{iballesterllagaria@gmail.com}

\author[1]{\fnm{Georg} \sur{Kosakowski}}\email{georg.kosakowski@psi.ch}

\author[4,5]{\fnm{Raphael} \sur{Wüst}}\email{raphael.wuest@nagra.ch}

\author[1]{\fnm{Nikolaos I.} \sur{Prasianakis}}\email{nikolaos.prasianakis@psi.ch}

\affil*[1]{\orgdiv{Laboratory for Waste Management}, \orgname{Paul Scherrer Institute}, \orgaddress{\street{Forschungsstrasse 111}, \city{Villigen PSI}, \postcode{5232}, \country{Switzerland}}}

\affil[2]{\orgdiv{Institute of Geological Sciences}, \orgname{University of Bern}, \orgaddress{\street{Baltzerstrasse 1+3}, \city{Bern}, \postcode{3012},  \country{Switzerland}}}

\affil[3]{\orgname{ETH Zürich}, \orgaddress{\street{Rämistrasse 101}, \city{Zürich}, \postcode{8092}, \country{Switzerland}}}

\affil[4]{\orgname{Nagra}, \orgaddress{\street{Hardstrasse 73}, \city{Wettingen}, \postcode{5430}, \country{Switzerland}}}

\affil[5]{\orgname{Earth and Environmental Science, James Cook University}, \orgaddress{\city{Townsville}, \postcode{4811}, \country{Australia}}}

\abstract{
Deep subsurface exploration is important for mining, oil and gas industries, as well as in the assessment of geological units for the disposal of chemical or nuclear waste, or the viability of geothermal energy systems. Typically, detailed examinations of subsurface formations or units are performed on cuttings or core materials extracted during drilling campaigns, as well as on geophysical borehole data, which provide detailed information about the petrophysical properties of the rocks.

Depending on the volume of rock samples and the analytical program, the laboratory analysis and diagnostics can be very time-consuming. This study investigates the potential of utilizing machine learning, specifically convolutional neural networks (CNN), to assess the lithology and mineral content solely from analysis of drill core images, aiming to support and expedite the subsurface geological exploration. The paper outlines a comprehensive methodology, encompassing data preprocessing, machine learning methods, and transfer learning techniques. The outcome reveals a remarkable 96.7\% accuracy in the classification of drill core segments into distinct formation classes. Furthermore, a CNN model was trained for the evaluation of mineral content using a learning data set from multidimensional log analysis data (silicate, total clay, carbonate). When benchmarked against laboratory XRD measurements on samples from the cores, both the advanced multidimensional log analysis model and the neural network approach developed here provide equally good performance. This work demonstrates that deep learning and particularly transfer learning can support extracting petrophysical properties, including mineral content and formation classification, from drill core images, thus offering a road map for enhancing model performance and data set quality in image-based analysis of drill cores.}

\keywords{Convolutional Neural Networks, Transfer Learning, Core Analysis, Lithology, Mineral Composition}

\maketitle

\section{Motivation and Background}\label{sec1}

Geological exploration of the underground is important for the mining, mineral, oil, and gas industries. In Switzerland, geological exploration of the underground for the deep geological disposal of radioactive waste is currently in progress and includes remote sensing (seismic data analysis), geophysical surveys (log analysis) and drill core laboratory analysis. Although the combination of these techniques provides precise and reliable results, laboratory investigations are often labor-intensive, time-consuming, and costly when large sample numbers and volumes are present. Combining conventional field and laboratory analytical techniques with machine learning may help enhance data analysis and provide a deeper insight into data \cite{jooshaki_systematic_2021,jung_systematic_2021,woodhead_harnessing_2021}.

In this paper, we demonstrate the capacity of machine learning respectively, deep learning, particularly emphasizing transfer learning to extract selected petrophysical properties, like mineral content or sample formation from drill core images.

Machine learning has been successfully used in the past to classify the lithology based on drill core images: Many of these studies applied convolutional neural networks (CNN), a special type of neural network that has proved to be successful for handling images. Particularly pretrained CNN architectures such as VGG16, DenseNet, ResNet, ResNest, or ResNext applied to  drill core samples from Norway, South Australia, Gulf of Mexico \& North Sea, China and Switzerland achieved good results for lithology classification with accuracies for the test data sets ranging from 60\% up to 99.6\% \cite{ alguliyev_automatic_2023,
 classification_lithology_3classes, falivene_lithofacies_2022,classification_lithology,lithological_hetorogeneity}. Furthermore, in this context, autoencoders \cite{park_assessment_2023, solum_accelerating_2022} and vision transformer architectures \cite{koeshidayatullah_faciesvit_2022} have also been applied for this task. These techniques were utilized for several thousand image slices from drill cores from  Western Australia, Gulf of Mexico \& North Sea and Russia. The testing accuracy varied from 70\% up to 96.4\%, and showed a strong dependency on the heterogeneity of the images and the number of classes that the samples were categorized. 

The analysis of the model performances reveals that the sample preparation and data preprocessing are paramount steps when applying machine learning methods. The necessary steps include the automatic detection of trays and cores \cite{BARABOSHKIN2022105099, gunther_towards_2021}, the assessment of rock quality, the classification of intact and non-intact cores, as well as the recognition of empty tray areas and non-rock objects \cite{alzubaidi_automated_2022, li_automatic_2023}. Depending on the application, for preprocessing the automated crack detection \cite{alzubaidi_automated_2022} can also be important. Moreover, the classification of lithology is done based on a narrow core interval and not all cores have the same length, thus the preprocessing step includes slicing the available data set into smaller images, ranging from 0.5x0.5cm up to 10x10cm, as used in the aforementioned works. Besides full drill core images, thin section images \cite{faria_lithology_2022, zhou_rock_2023}, rock sample images \cite{shi_novel_2023}, or rock debris \cite{xu_multi-feature_2022} images were used to perform rock type classification. Some approaches to predict lithology from drill core images involve additional data to increase accuracy, like elemental information (\cite{xu_integrated_2021, xu_intelligent_2022}), petrophysical and geochemical data (\cite{houshmand_rock_2022}), or pXRF measurements (\cite{trott_random_2022}), and the use of machine learning models designed for data fusion. 

Assessing the mineral content has so far only been done with different types of data, not drill core images alone. In \cite{tusa_drill-core_2020}, mineral abundance predictions relied on a combination of hyperspectral short-wave infrared data and, for small areas, additional Scanning Electron Microscopy-based images with different machine learning approaches: Random Forest (RF), Support Vector Machine (SVM), and Neural Networks (NN). A model based on RF was developed in \cite{quantitative_mineral_mapping} that estimated the mineral proportion from Long-wave infrared spectra. For labeling the training data, micro-X-ray fluorescence measurements were used. Spectral and geochemical data were utilized to train a CNN model that predicts the Cu concentration in \cite{guerra_prado_ore-grade_2023}. In \cite{krupnik_high-resolution_2020, kupssinsku_hyperspectral_2022-1}, hyperspectral data combined with different machine learning approaches like SVM, NN, and Spectral Angle Mapper were investigated for mineral mapping and porosity estimation. Also, spectral data, in terms of multi-sensor spectral imaging, together with SVM, were used to distinguish between six mineralogically meaningful classes, and the corresponding probability estimates of each class were derived in \cite{multi_sensor}. In the study discussed in  \cite{mishra_irida_2022}, core plug samples were combined with continuous Kimeleon colorlith logs, which use information from the apparent matrix density, neutron porosity, and gamma-ray logs. K-means clustering was then used to classify the different rock types. From these continuous rock types, rock properties like permeability were calculated and up-scaled.

The goal of this work is to train neural networks that take only drill core images as input and can assess the lithological classes and the mineral content, respectively. We first perform a simplified lithological classification of drill core images from Northern Switzerland into distinct geological formations using pretrained CNNs designed for image classification.
Inspired by the recent progress in automated image processing, we analyze the core images stemming from deep drilling exploration of the Mesozoic underground conducted by Nagra, the Swiss National Cooperative for the Disposal of Radioactive Waste \cite{NAGRA}, in the context of the national program on site selection for Swiss deep geological repository for radioactive waste \cite{NagraTru11}.

The second step and principle objective of this research is to assess the performance of neural networks in predicting the mineral content (amount of carbonate, silicate and total clay) from solely the drill core images.
Because none of those studies mentioned above considers the drill core images alone to predict the mineral content.
For this purpose, parts of the classification model are used via transfer learning. Thereby, even with a relatively limited dataset (in our case 361 data points) the mineral content regression from drill core images is possible. For training the CNN, the images are labeled with mineral content data retrieved from a multidimensional log analysis of the petrophysical logs from one borehole. The so trained model is then applied to previously unseen drill core images of the same borehole and the predictions are compared to bulk XRD measurements of the mineral content.

The paper is structured in the following way: First, the available data set, consisting of the drill core images and additional measurements for labeling, is described in detail (Section \ref{dataset}). Section \ref{methodology} introduces the methodology, covering data preprocessing, machine learning methods, and transfer learning. Results for the formation classification and mineral content regression are presented in Section \ref{results} as well as details on the comparison of the regression model predictions to measured mineral contents. This is followed by final remarks and outlook for further research investigations in Section \ref{conclusion}. This study not only advances geological analysis but also underscores the potential of machine learning to enhance subsurface exploration and characterization.

\section{Data set}\label{dataset}

\begin{figure}[h]
    \centering
    \includegraphics[width=\textwidth]{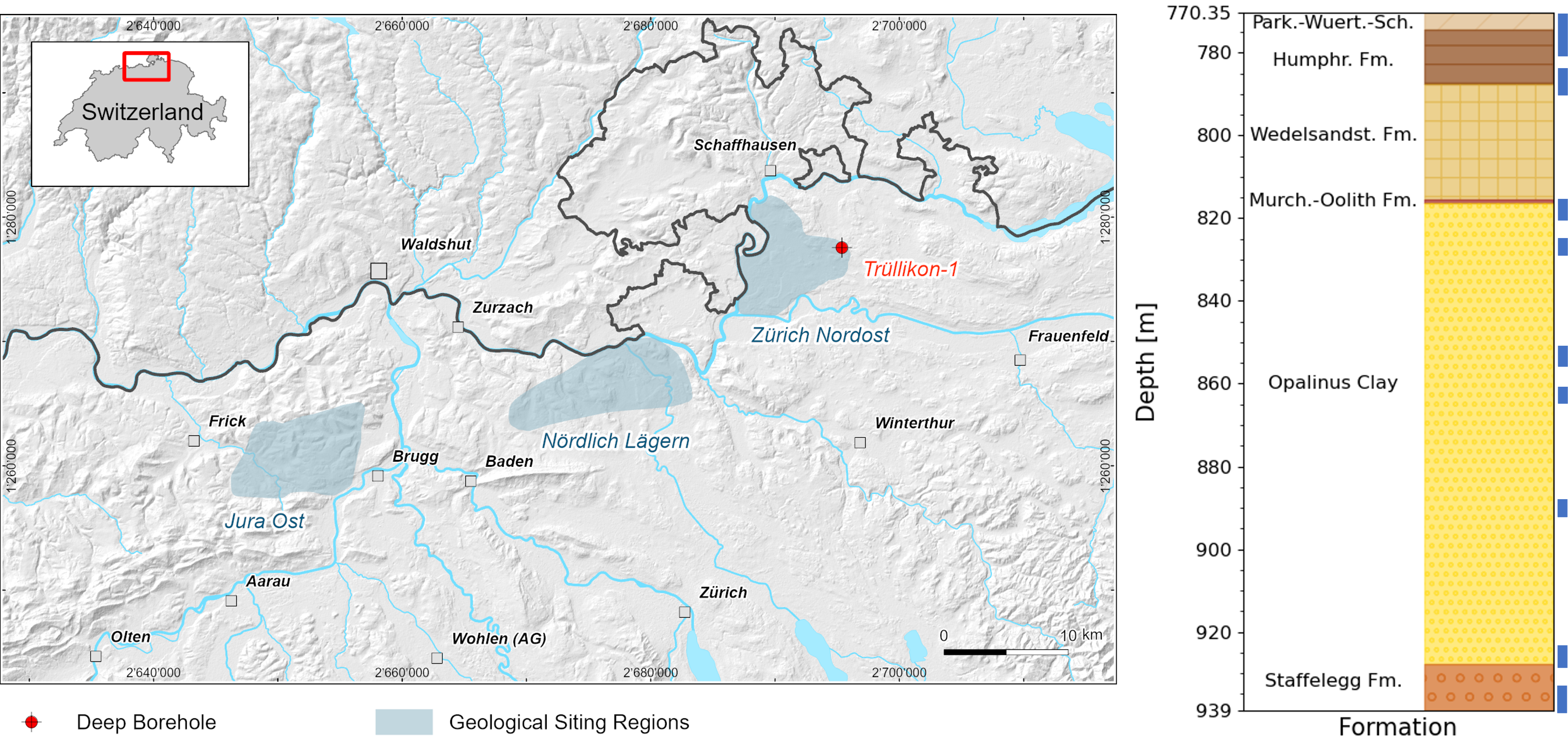}
    \caption{Left: Cartographic representation illustrating the geographical region of northern Switzerland. The map specifically features the borehole site of Trüllikon 1-1 which is selected for the data analysis of this study.  Background: ©Data:swisstopo and hillshading from NASA SRTM. Right: Abstracted lithological profile of the section Trüllikon 1-1 showing the interval between 770.35m and 939m depth. The blue bars denote the available data segments.}
    \label{fig:standort}
\end{figure}

The samples and data used in this study are from the borehole Trüllikon 1-1 in the siting region of Zürich Nordost (left Figure \ref{fig:standort}) in Switzerland, \cite{NAGRA}. The borehole is part of Nagra's national program on site selection for Swiss deep geological repository for radioactive waste, \cite{NagraTru11}. The total drilling depth was 1310m. The section between 498-1029m was cored with wireline coring. At the time of the study, only a fraction of the cored section was measured, validated, and released, comprising non-continuous segments of in total 55m of core between 770.35m and 939m depth of sedimentary rocks. The available data segments are visualized by the blue bars next to the lithological profile shown in Figure \ref{fig:standort}, right. Detailed information on all drilling procedures and data can be found at \cite{NAB20_09, NAB20_09_I}. 

The extracted drill cores were washed, dried and photographed using a standardized procedure together with a ruler and reference cards (X-Rite ColorChecker Classic Mini color chart and BST14 gray scale chart) for samples referencing and colour calibration. For the high-resolution (10px/mm) core photographs, a DMT CoreScan system from avaluar GmbH with a CCD-Colour camera CSc3b1.26 was used. Note that only single core images, not the 360 degree photos were considered in this work. Further details on the photographing conditions can be found in \cite{NAB20_09_II}. The photographs were stored along with the corresponding drilling depth. An example of such photographs is shown in Figure \ref{fig:preprocessing} (a). 

Along with the photographs, data from the petrophysical Multimineral Log Analysis (MultiMin) \cite{NAB20_09_X} was used for the labeling in the regression task. 
The MultiMin method is a stochastic workflow for log analysis based on the assumption that every borehole log measurement is determined by the mineral and fluid content of the rock that surrounds the borehole. If the linear or non-linear relations between measured properties and rock composition are known, it is possible to calculate the theoretical log response for a given rock composition or use all available log information to estimate the rock properties of interest. By comparing measured and predicted log values, it is possible to assess the quality of the estimated properties. For details see Chapter 3.3 in \cite{NAB20_30}.  
Petrophysical logs available for the analysis included caliper log, gamma ray log, spectral gamma ray log for U, Th and K, neutron hydrogen index,  gamma-gamma density log, element spectroscopy, electrical resistivity, and a sonic log. In addition, multi-sensor core logger data (bulk density, compressional (P) wave velocity, spectral gamma ray curves for K, Th and U, and X-ray fluorescence elemental analysis for Fe, Si, Ca, Al, Ti and S were available for Opalinus Clay and its confining units. These logs and lab measurements of mineralogy (XRD), porosity and density were used for the MultiMin modeling of porosity and mineral composition; details on that can be found in the Nagra working reports \cite{NAB20_09_X, NAB20_30}. 

The MultiMin model was utilized to calculate mineral composition approximately every 15 cm in terms of clay minerals (kaolinite, illite, smectite, and chlorites), other (alumino) silicates (quartz, potassic feldspars, plagioclase - this group of minerals will be referred to as silicates for the rest of this work), carbonates (calcite, siderite, dolomite, ankerite), iron oxide, evaporates (anhydrite), and organic carbon (kerogen). For the sake of simplicity and robustness of the CNN model, only the total amount of total clay, carbonate, and silicate was considered for the regression. 

The core samples are grouped to six distinct formations:  
1. Parkinsoni-Württembergica-Schichten (738.97-774.55 m), 
2. Humphriesioolith Formation (774.55-787.50 m), 
3. Wedelsandstein Formation (787.50-815.51 m), 
4. Murchisonae-Oolith Formation (815.51-816.42 m), 
5. Opalinus Clay (816.42-927.91 m), 
6. Staffelegg Formation (927.91-971.68 m);
with the numbers in brackets giving the measured core depth.
A visual representation of the abstracted lithology is provided in Figure \ref{fig:standort}, right. 

For validating the trained neural network model, bulk XRD measurements were used. Within the 55 meters of core, a total of 23 bulk XRD measurements were available. The exact positions within the core are detailed in Table \ref{tab:meas_model}, Appendix \ref{App:XRD}, and are visually accessible in the core photograph overview in the Appendix \ref{App:Corephoto} (highlighted by blue rectangles.) 

\section{Methodology}\label{methodology}

\subsection{Image Data Preprocessing}

To facilitate robust and automated analysis of drill core photographs, a standardization process for colour balance and image dimensions was performed, and artifacts, such as cracks or missing fragments were labeled. 
Accordingly, a five-step preprocessing routine was introduced for all images. 
First, the photographs have been corrected for color (Step 1) and white balance (Step 2). Then, an automated algorithm for core segmentation from background (Step 3), core splitting into equally sized segments (Step 4), and crack detection (Step 5) was applied. 

The preprocessing was implemented using the Python Imaging Libraries PILLOW and scikit-learn, \cite{clark2015pillow, scikit-learn}. The five distinct steps of the automatic preprocessing pipeline are described in detail below and are as follows:

1. The first step of the color correction involved creating a color profile (ICC profile) using the ColorChecker from X-rite library. The ICC profile was then subsequently applied to the images. This step ensured that the color representation across the data set was consistent throughout the images.

\begin{figure}[h]
    \centering
    \includegraphics[width=\textwidth]{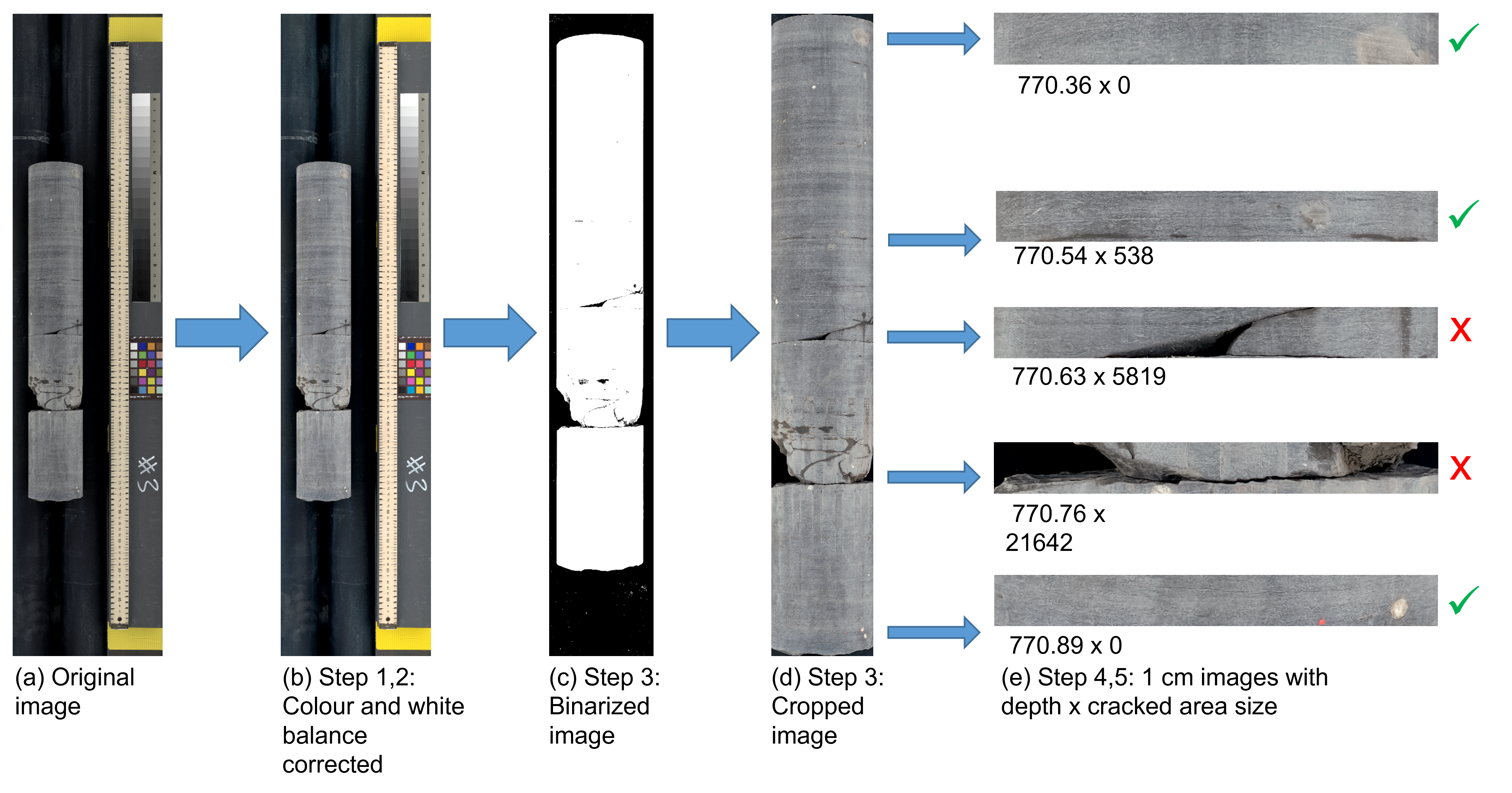}
    \caption{Preprocessing pipeline, consisting of five distinct steps, from the original image (a) to the final 1cm image segments (e).  The 1cm segments with the red cross next to the image are excluded from the analysis, since the cracked area size is bigger than 5000 pixels.}
    \label{fig:preprocessing}
\end{figure}

2. The second step implemented the correction of the white balance using the white colour patch (A) as a ground truth for true white. By comparing the color values of the white patch with the rest of the image, the pixel values of the image were normalized relative to the maximum value found on the white patch. Figure \ref{fig:preprocessing} (a) shows one example from the data set. The color and white balance corrected image is depicted in Figure \ref{fig:preprocessing} (b).

3. The third step consisted of subtracting the background and extracting the domain representing the drill core from the photograph (Figure \ref{fig:preprocessing} (c), (d)). The segmentation algorithm first used the length and depth of the core piece, recorded in the name of the photograph, to approximately locate and cut the core domain from the background (Figure \ref{fig:preprocessing} (c)). This was possible, since the core was always centered in the middle of the image. Further, Otsu's method (\cite{otsu1979}) was used to separate the core from the background based on intensity distribution. Morphological operations were then applied to refine the segmentation. Next, continuous regions that belong to the core were identified and used to create a bounding box that accurately encompasses the core. The image was subsequently cropped using this bounding box, focusing the analysis exclusively on the core region for further processing and evaluation (Figure \ref{fig:preprocessing} (d)).

4. In the fourth step of the preprocessing pipeline, the extracted core images were divided into equally sized 1 cm depth segments. The segmentation algorithm was based on the core length obtained from the image filename, resulting in images with a width of around 100 pixels and a height of 850 pixels corresponding to a 1 cm segment of the drill core with a certain depth. (Figure \ref{fig:preprocessing}.(e)). If a segment overlapped two continuous core images, they were merged to form a cohesive segment. This division resulted in 5496 core segments from all available photos, out of which 361 could be used for creating the regression model to predict the mineral content from the 1 cm drill core images. The small number of images for the regression task is due to the number of corresponding data from the MultiMin analysis. The remaining core segments (i.e. 5135) were used for the classification task.

5. In the final, fifth, step, cracked segments were detected by binarizing the image again with Otsu's method. The images containing a crack were not excluded from the beginning, but the size of the cracked area per segment was stored in the filename. The segment was saved as a TIFF file with the naming convention: “DxC.tif,” where D was the depth and C was the cracked area size in pixels (Figure \ref{fig:preprocessing} (e)). These images provided the input for the machine learning models. Images, where the cracked area size exceeded a certain threshold of, e.g., 5000 pixels, were removed from the data set for data quality improvements and thus better model performance. % to improve the quality of the data and therefore increase the performance of the models.

\subsection{Neural Networks and Transfer Learning}
Neural networks have emerged as powerful tools for modeling a functional relationship between generic input and output. In particular, convolutional neural networks (CNN) have been proven specifically useful in image analysis, as they can autonomously learn feature engineering through the use of filters. Here, the basic principles of CNN usage are described, without providing intricate details, those can be found in \cite{10.5555/3378999}.

Two neural network architectures were applied in this work. The first NN architecture was designed and trained to take the 1cm drill core images as an input and predict the formation class as an output (i.e. classification task). The second NN architecture was intended for regression, and it was set up to take the 1cm drill core images as an input, and estimate the content of clay, carbonate and silicate minerals in the core as output. 

\begin{figure}[h]
    \centering
    \includegraphics[width=\textwidth]{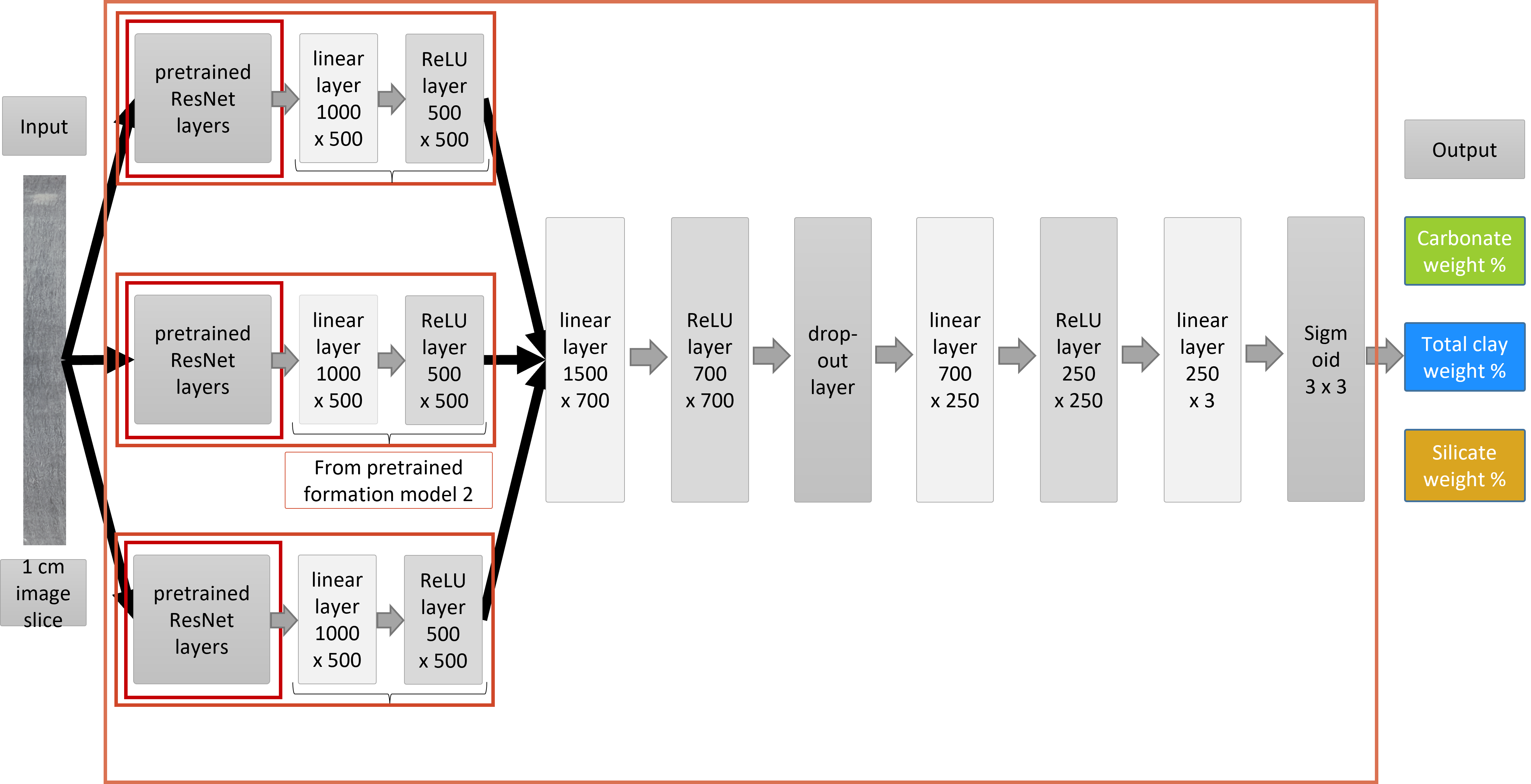}
    \caption{Architecture of the neural network for the mineral content regression with three formation models as a backbone architecture}
    \label{fig:CNN-architecture2}
\end{figure}

For that, the concept of transfer learning was used rather than creating and training a neural network architecture from scratch. In this process, a model trained for a similar but different task served as a backbone architecture and additional layers were placed on top of the pretrained layers, see Figure \ref{fig:CNN-architecture2}. Training such a neural network can be done in two ways, either all weights and biases are learned, or the ones of the pretrained model are fixed and only the new ones are updated. 

In this work, pretrained deep convolutional neural networks (VGG16, ResNet18, ResNet34, ResNet50, ResNet101, ResNet152 \cite{7780459}) were used to generate appropriate models. The residual neural networks (ResNet) are particularly powerful because compared to very deep convolutional neural networks (like VGG) they overcome the vanishing gradient problem by using shortcuts (skip-connections). 
The classification and regression models using the different pretrained convolutional neural networks were implemented using pytorch, \cite{NEURIPS2019_9015}, a machine learning library for python and C++. The pretrained models implemented in pytorch, were originally trained on the ImageNet data set, storing their weights and biases. The ImageNet data set comprises over 14 million hand-annotated images, spanning thousands of categories, such as vehicles, animals, persons, fruits, and geological formations. 
For our models, pretrained models on this data set are responsible for extracting general features from images. Therefore the weights and biases of these pretrained backbone architectures were kept fixed during the training process of our specific models. While the last layers added on top of the pretrained models were adapted for the specific tasks. This is the common state-of-the-art procedure in image classification applications. 

For training neural networks, typically the loss function is minimized, which computes the difference between the actual and predicted output targets. Different optimization algorithms are available. In this work, the Adam optimization algorithm was used together with a cross-entropy loss for the classification and a mean squared error loss for the regression.
The training is additionally controlled by an early stopping procedure. This entails continuing training as long as the loss is decreasing and the performance on the validation data set is increasing. If this is not the case for a few epochs, then the training is stopped.
The metrics used for evaluating the model are the accuracy for the formation classification and the (root) mean squared error and coefficient of determination R2 for the mineral content regression. 
To see the generalization performance of the neural network and to not introduce a bias through transfer learning, the data set was divided as follows: 5135 image segments were used solely for classification, and another 361 (i.e. 6.6\%) solely for regression. For both tasks, the data were subdivided into three sets, the training and validation data set used for training, and the test data set used solely for testing after the training had been completed. 
More specifically for the classification task with a crack threshold below 5000 pixels, 4658 images were used for training (Parkinsoni-Württembergica-Schichten: 330 images, Humphriesioolith Formation: 911 images, Wedelsandstein Formation: 250 images, Murchisonae-Oolith Formation: 43 images, Opalinus Clay: 2498 images, Staffelegg Formation: 626 images), 114 images for validation (19 images per class) and 113 images for testing (Parkinsoni-Württembergica-Schichten: 18 images and 19 images in all other classes). For the regression task, 276 images were used for training, 34 images for validation and 34 images for testing in the case of a crack threshold of 5000; for the threshold of 1000, it was 254, 32 and 31 images for training, validation and testing. Among the images for the regression task, in all images there were mixtures of clay, carbonate and silicate and not purely one element.  It is important to mention that the images used for the regression task are independent from the images used for the classification task. So no images were used twice in the whole process. Also all the training, validation and testing images were independent from each other. All the images amongst many others can be found in \cite{NAB20_09_II}, in this document the cores are extracted from the background. 

\section{Results and Discussion}\label{results}
\subsection{Formation Classification}\label{classification}

Six different pretrained backbone architectures (ResNet18, ResNet34, ResNet50, ResNet101, ResNet152, VGG16) were benchmarked for the formation classification task. 
The ResNet architecture requires same size and normalized input images. 
To use the image segments as input for the ResNet, the segments were loaded into a range between 0 and 1,  normalized by using the mean and standard deviation, and resized to 850x100 pixels. The output of the ResNet was then a 1000-dimensional vector.  During training, the loaded weights and biases of the pretrained network were fixed. The 1000-dimensional vector returned by the backbone architecture served as input for the classification part. For the formation classification, the pretrained neural network was complemented with a linear layer of input 1000 and output 500 dimensions (1000x500), a ReLU layer (500x500) and again a linear layer (500x6). The last layer had 6 output dimensions due to the chosen encoding. Since the formation classes were categorical, each class needed to be encoded to a number; this was done with one-hot encoding, where the first class was represented as (1,0,0,0,0,0), the second as (0,1,0,0,0,0), and so forth until the last class (0,0,0,0,0,1). The implemented and trained model gave in the end the number of the predicted class, as an integer among $\{0,1,2,3,4,5\}$.

\begin{figure}[h]
    \centering
    \includegraphics[width = \textwidth]{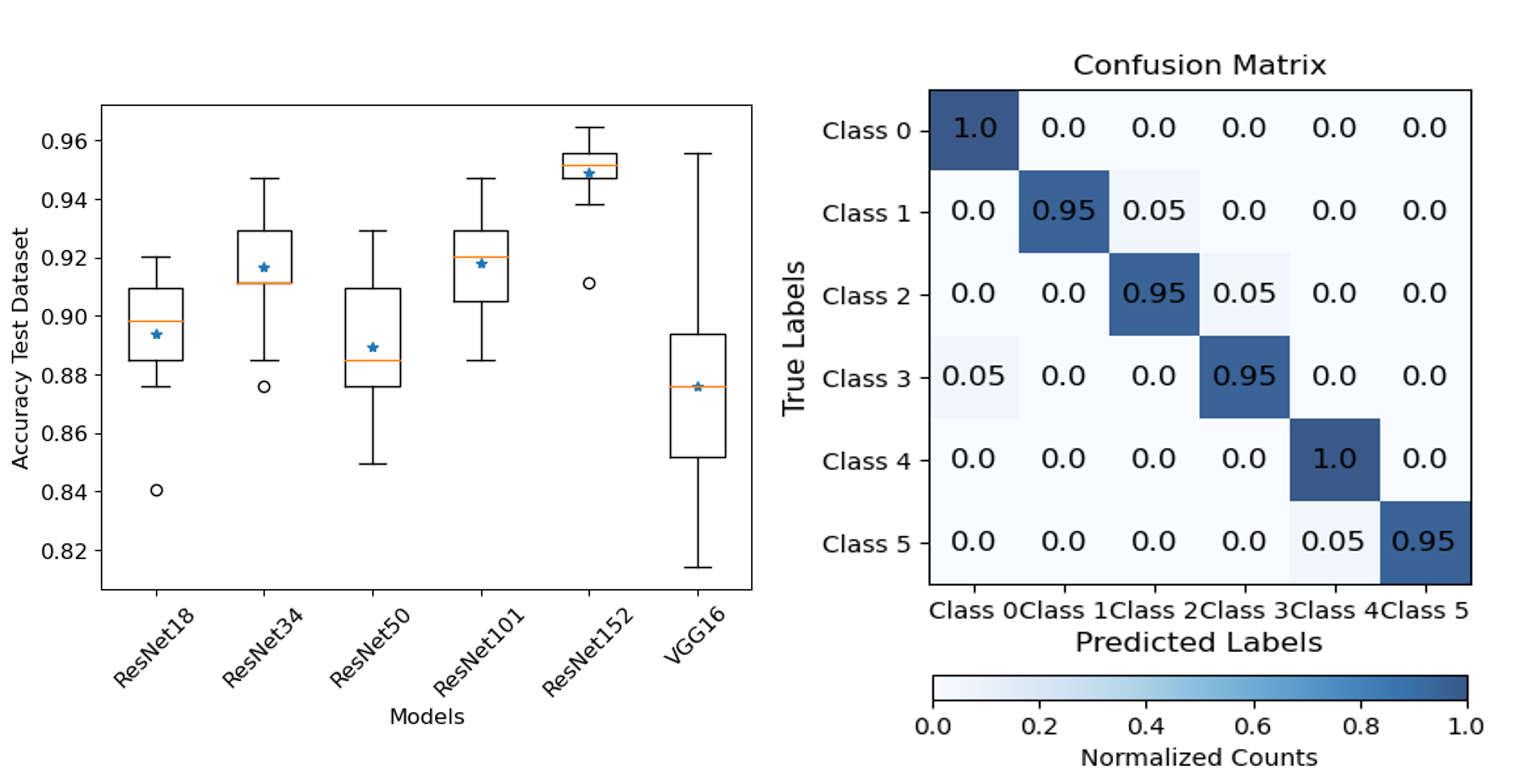}
    \caption{Left: Formation classification results for the test data set for different backbone architectures with 14 different seeds each. Right:  Confusion matrix for the formation classification model based on ResNet152 with artefact threshold T=5000 pixels for the test data set.} \label{figboxplot}
\end{figure}

\begin{figure}[h]
    \centering
    \includegraphics[width=\textwidth]{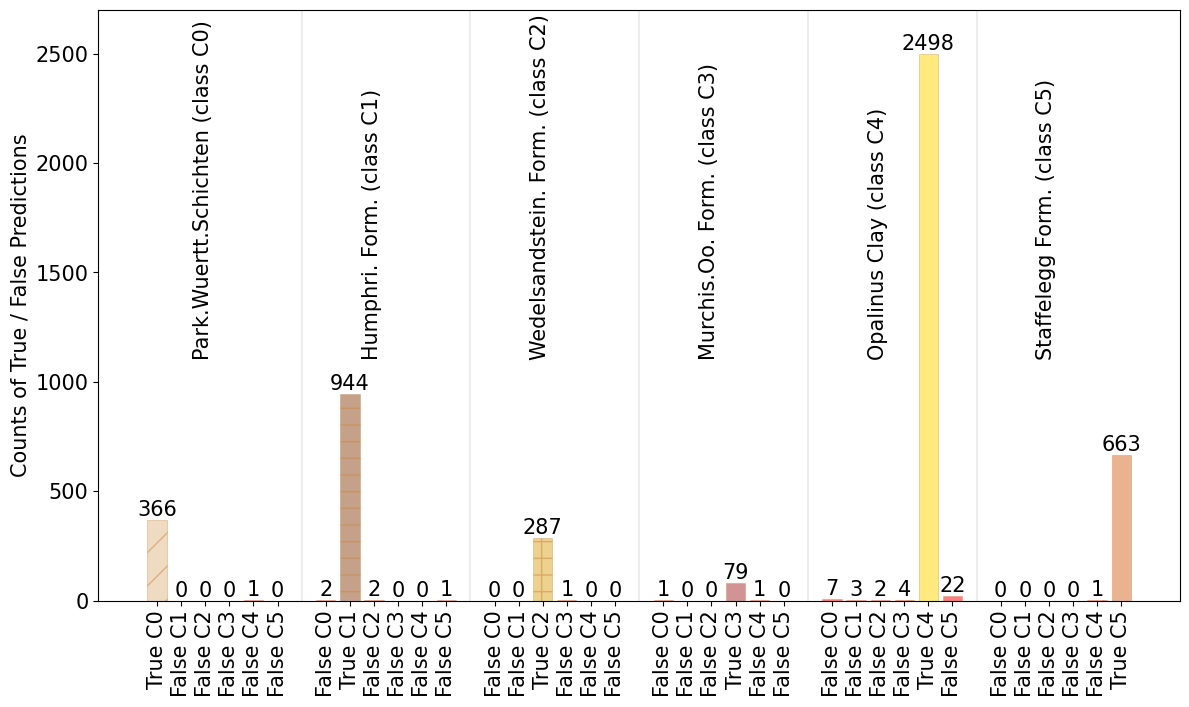}
    \caption{Counts of true and false predicted classes for each formation. In the x-axis, the classes are numbered; C0 corresponds to Park.Wuertt.Schichten, C1 to Humphr. Formation, ..., C5 to Staffelegg Formation. For a better understanding of the figure, consider e.g. the Park. Wuertt.Schichten, here the first 6 x-labels are relevant, the number of counts, 366 for True C0 means that 366 images were correctly classified as Park. Wuertt.Schichten, so class C0; no image was classified as class C1, C2, C3, or C5 and only one image that should have been a Park. Wuertt.Schicht was classified as class C4 (Opalinus clay), so this was misclassified.  }
    \label{fig:barplot_extended}
\end{figure}

After the training phase, the performance of the models was evaluated exclusively with the test data set in terms of accuracy. The graphical representation of the results, depicted as box plots derived from 14 different seeds for training initialization, is presented in Figure \ref{figboxplot}(a). In these plots, the blue star denotes the mean, the horizontal orange line represents the median, while the box encapsulates the lower and upper quartiles. The "whiskers," short horizontal lines, denote the values for the first (lower) and third (upper) quartiles $ \pm 1.5 $ times the interquartile range, which is the length between the first and third quartiles. Outliers are marked as individual circles. As can be seen in Figure \ref{figboxplot}(a), the highest performing models are the ones with ResNet152 as a backbone architecture with the best one having an accuracy of 96.7\%. For the overall best model, the confusion matrix is depicted in Figure \ref{figboxplot}(b). This matrix illustrates the comparison between the predicted and true formation classes for the test data set. Considering, for example the true class 2 (i.e. Wedelsandstein Formation) in row 3, one can see in the matrix that 95\% of the images were classified correctly as class 2, whereas 5\% were wrongly classified by the model as class 3. The plot reveals that misclassified images primarily correspond to neighboring classes. To investigate this further, the analysis was extended to the whole data set. 
Notably, out of 4885 images, only 48 were falsely classified (1 \%), with the majority of these misclassifications occurring between adjacent classes. 
Figure \ref{fig:barplot_extended} illustrates the number of true and false predicted image slices for each formation.
Among the 48 falsely predicted image slices, 8 images between 891.03 to 891.18m depth were predicted to fall into the Staffelegg formation instead of the Opalinus Clay. This misclassification is attributed to the high lithological similarity between the Opalinus Clay and the Staffelegg clay-rich rocks.

Similarly, 14 images were misclassified at the boundary between these two formations which again is attributed to little lithological difference between the samples at the boundary. Therefore, it is important to acknowledge that for classification performance of the CNN on core samples, stark differences between rock compositions facilitate differentiation, while similar or weak differences may lead to false classifications and further details would need to be considered.

The accuracy achieved at this stage is considered satisfactory and no further refinement of the models was pursued. 

\subsection{Mineral Content Regression}\label{regression}
\subsubsection{Comparison of Neural Network Model Architectures}
Initially, the same architecture as for the formation classification, but with a linear layer of output dimension 3 was tested for the mineral content regression. The performance of this model was very poor with an average R2 value of -0.23 over 15 different seeds. Hence, a different strategy was developed for the mineral content regression. Since the formation classification model performed very well, the trained formation classification neural network, except the last layer, was used as the backbone architecture for the regression model. A linear layer (500x250), a ReLU layer (250x250), followed by a dropout layer and finally a linear layer (250x3) were added on top to solve the regression task (see Fig. \ref{fig:CNN-architecture} in the appendix \ref{App:regression}). With such an architecture, the classification of the images according to the rock type is performed first and helps to perform mineral composition analysis. The same strategy was also tested not only with one but with the three formation classification models showing the best performance, which were concatenated, and the following layers were added on top, see Fig. \ref{fig:CNN-architecture2}: A linear layer (1500x700), a ReLU layer (700x700), a dropout layer, a linear layer (700x250), a ReLU layer (250x250), a linear layer (250x3), and a sigmoid layer (3x3). Additionally, models were trained using a reduced crack detection/deletion threshold of 1000 pixels, as opposed to the previous threshold of 5000 pixels, for the model employing the ensemble of three formation classification models.

In summary, three model types were trained and tested for mineral content regression. These were: 1) the best formation classification model as backbone architecture with crack detection/deletion threshold of 5000 pixels (T5000-1m); 2) the best trio of formation classification models as backbone architecture with a threshold of 5000 (T5000-3m); 3) the best trio of formation classification models as backbone architecture with a threshold of 1000 (T1000-3m).

Each model type was trained 18 times with different seeds. The outcomes in terms of the coefficient of determination (R2) for the test data set are reported in Table \ref{tab:Mean_max_R2} and are visualized in Figure \ref{fig:boxplot_regression_appendix} in the Appendix \ref{App:regression}. Data interpretations need to be aware that data used for training are derived from a model with inherent uncertainty.

\begin{table}[h]
\begin{tabular}{@{}l|ll@{}}
\toprule
          & R2 test mean & R2 test max \\ \midrule
T5000\_1m & 0.560        & 0.641       \\
T5000\_3m & 0.617        & 0.691       \\ 
T1000\_3m & 0.580        & 0.673       \\
\end{tabular}
\caption{Mean and maximum R2 values for the test data set for the three different model types. }\label{tab:Mean_max_R2}
\end{table}

\begin{figure}[h]
    \centering
    \includegraphics[width=\textwidth]{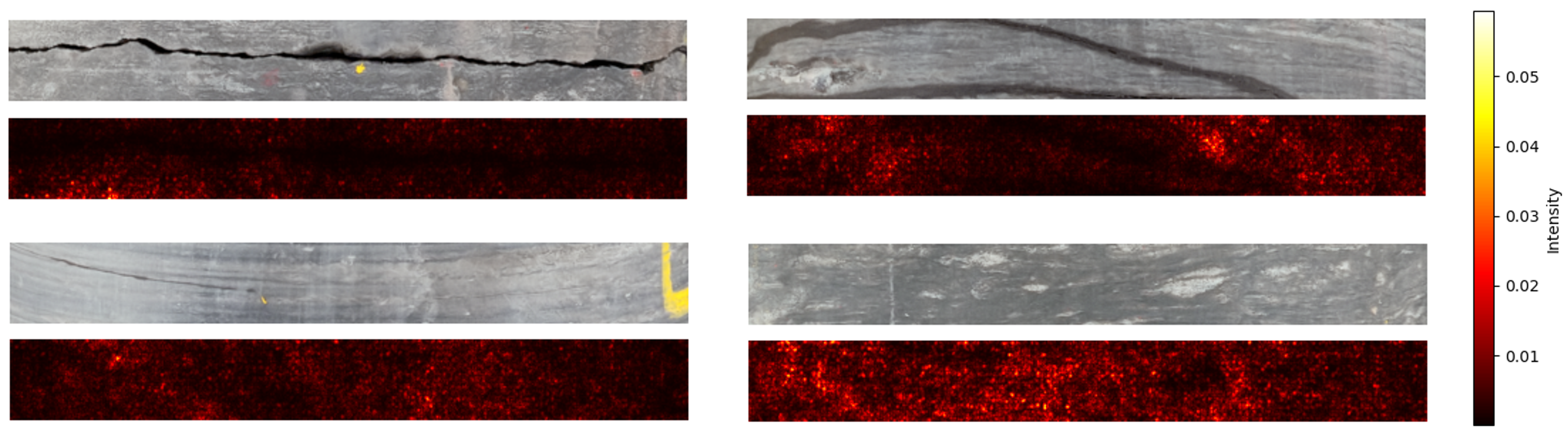}
    \caption{Saliency maps of 4 different 1cm image slices, upper images show the true image, lower images the corresponding saliency maps}
    \label{fig:saliency}
\end{figure}

The data in Table \ref{tab:Mean_max_R2} suggest a subtle better performance in the regression models that utilize the ensemble of three formation models as a backbone architecture. The R2 values for the models with a larger crack deletion area of 5000 pixels are higher, but further investigations are needed to determine the significance of detected cracks on the mineral content regression modeling. Therefore, saliency maps were considered, to better understand the importance of crack detection and further need to dismiss these images. A saliency map for CNNs highlights the most relevant regions or features within an input image, helping to identify areas of significance for the particular task. The assigned dark colors refer to the "unimportant" domains, whereas the lighter red domains are the most "significant" ones for the neural network. 
From these maps, as depicted in Figure \ref{fig:saliency}, one can see that cracks or color marks appear very dark in the saliency maps, meaning that those features are not so relevant for solving the regression task. Hence, the size of the crack deletion area was not considered further within this work.  

\subsubsection{Performance of Best Mineral Content Regression Model}
\begin{figure}
\includegraphics[width = \textwidth]{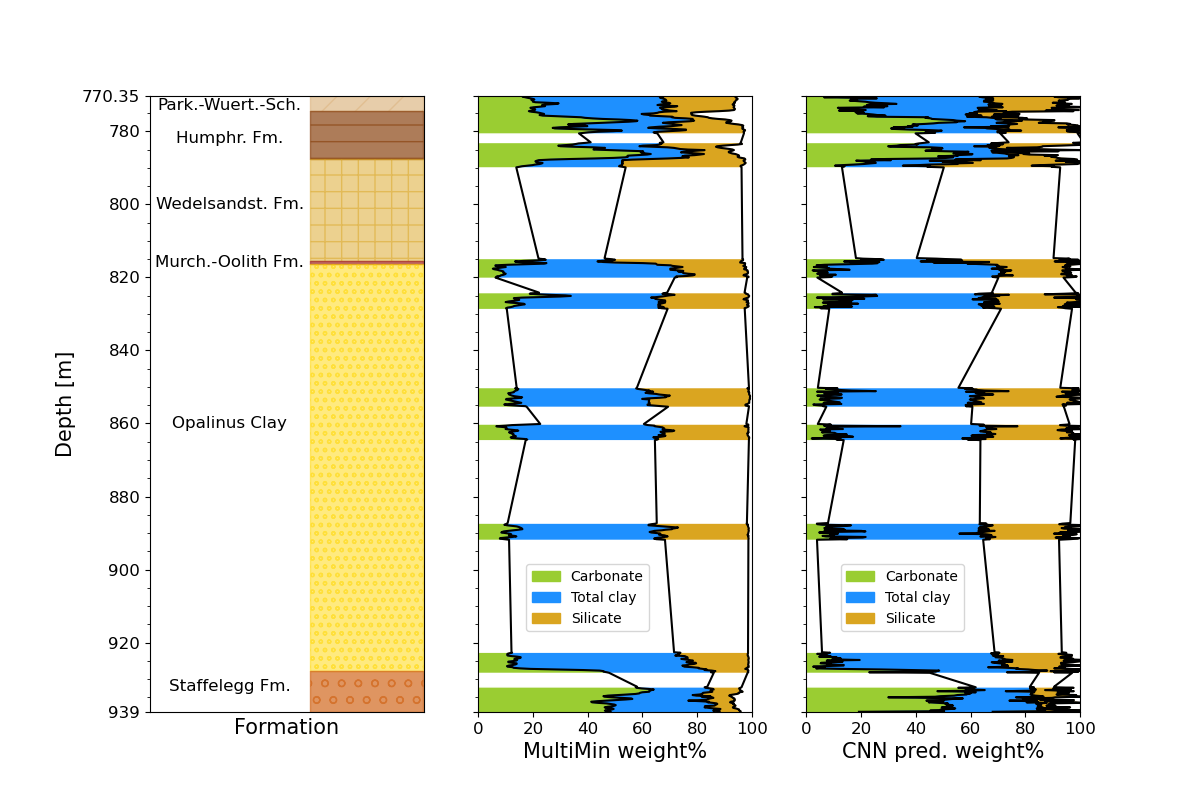}
\caption{ The first column shows the various formations along the depth, the second shows the predictions of mineral content with MultiMin model, whereas the third column shows the CNN predicted mineral content in this rock. Note that images were only available for the coloured areas. }\label{fig:regression_depth}
\end{figure}

The best model according to the R2 values on the test data set was the one with three formation models as a backbone architecture and a threshold of 5000 pixels and a seed of 1000. The comparison of the multiMin model data and the CNN predicted mineral content is depicted in Figure \ref{fig:regression_depth}.
For this best mineral content regression model, the absolute and relative errors as well as the R2 values for each individual mineral within the test data set were computed:
The model exhibits a moderate level of accuracy in predicting carbonate mineral content, with an absolute error of 0.059 and a relative error of 39.1\%. While the model's R2 value of 0.609 indicates a reasonable fit to the data, further refinement may be necessary to enhance the accuracy of the model. In contrast, the model performed much better in predicting the silicate mineral content, with an absolute error of 0.038 and a relative error of 18.7\%. The higher R2 value of 0.707 underscores the model's ability to explain the variance in silicate content. The best results with the trained models were obtained from the prediction of the total clay content, with an absolute error of 0.046 and a lower relative error of 10.7\%. The R2 value of 0.811 refers to a robust correlation that emphasizes the model's capability to capture the variability in total clay content effectively (See also Figure \ref{fig:regression_depth}).  
The high prediction accuracy for the total clay content may be due to several reasons including better spectral correlations with the clay phases. 

In Appendix \ref{App:XRD} Figure \ref{fig:meas_pred_single_mineral}, another direct comparison, of the MultiMin log data on the x-axis versus the predictions on the y-axis is given. 
The overall trend looks quite assuring and is supported by the metrics on the test data set, although there are some differences when considering all the details. Nevertheless, the model is already well suited for distinguishing between low and high mineral content for each species.

\subsubsection{Model Performance across Formations}
To investigate the model performance further, the absolute and relative errors for the mineral content predictions for each formation were investigated: The results are depicted in Table \ref{abs_rel_error_regression}, where nr gives the number of considered data points per formation. From the table, one can see that the performance varies a lot among the different formations. Summarizing, the lowest error values were achieved as follows: carbonates content was predicted best for the Humphriesioolith and the Staffelegg Formation with a relative error below 9.7 \%. Clay minerals content predictions showed a relative error below 6.8\% for the Parkinsoni-Württembergica-Schichten and the Opalinus Clay, while silicates predictions had a relative error below 5.1\% for the Wedelsandstein Formation. Throughout the analysis, across all formations a systematic better prediction emerged for the mineral phases, which were prevalent in the selected formations. 
One reason for the differences among the various formations could be, that the number of data points for training the model was not the same for each formation class. In addition to improving the results, assuming that many more data points may be available, one model for each formation could be trained.

\begin{table}[h]
\begin{tabular}{@{}r|rrrrrrr@{}}
\toprule
\textbf{Formation} & \textbf{nr} & \textbf{\begin{tabular}[c]{@{}r@{}}Carbonate\\ abs.err.\end{tabular}} & \textbf{\begin{tabular}[c]{@{}r@{}}Total clay\\ abs.err.\end{tabular}} & \textbf{\begin{tabular}[c]{@{}r@{}}Silicate\\ abs.err.\end{tabular}} & \textbf{\begin{tabular}[c]{@{}r@{}}Carbonate\\ rel.err.\end{tabular}} & \textbf{\begin{tabular}[c]{@{}r@{}}Total clay\\ rel.err.\end{tabular}} & \textbf{\begin{tabular}[c]{@{}r@{}}Silicate\\ rel.err.\end{tabular}} \\ \midrule
Park  & 26  & 0.03 & 0.025  & 0.025 
& 16.42\% & 5.21\% & 10.87\% \\ 
Humphr  & 69  & 0.037   & 0.037  & 0.037  & 9.62\% & 13.37\% & 19.86\% \\
Wedel  & 22  & 0.033  & 0.035   & 0.022   & 15.67\%  & 11.24\%  & 4.89\%  \\ 
Murch  & 5 & 0.033   & 0.038  & 0.067 & 21.06\%  & 11.17\%   & 16.00\% \\ 
Opa   & 173   & 0.051   & 0.036  & 0.027 & 41.14\%   & 6.73\%  & 10.17\%  \\ 
Staffel & 49  & 0.045   & 0.036  & 0.034 & 8.96\% & 10.75\%  & 36.62\%    \\ 
\end{tabular}
\caption{Comparison between the MultiMin log data and the CNN model prediction in terms of absolute (abs.err.) and relative (rel.err.) error of all three minerals for each formation. nr gives the number of data points available per formation.  }\label{abs_rel_error_regression}
\end{table}

\subsection{Comparison with bulk XRD measurements}
The training of the CNN for the mineralogical analysis was performed using data provided by MultiMin log model, which provides an indirect prediction of mineral content based on drill-logs data. The only measured data available for the true evaluation of the model's performance are the actual bulk XRD measurements performed on the cores. As explained in \cite{NAB20_09_X, NAB20_30} the core data of porosity, mineralogy and rock density were used for calibration of log measurements in the MultiMin workflow, although this does not mean that the MultiMin model data exactly reproduce the measured core data. The difference between modelled and measured data allows us to evaluate the performance of the model. Therefore, the analysis of the performance of CNN predictions and MultiMin log model were both performed against true XRD core measurements. 

\begin{figure}[h]
    \centering
    \includegraphics[width = \textwidth]{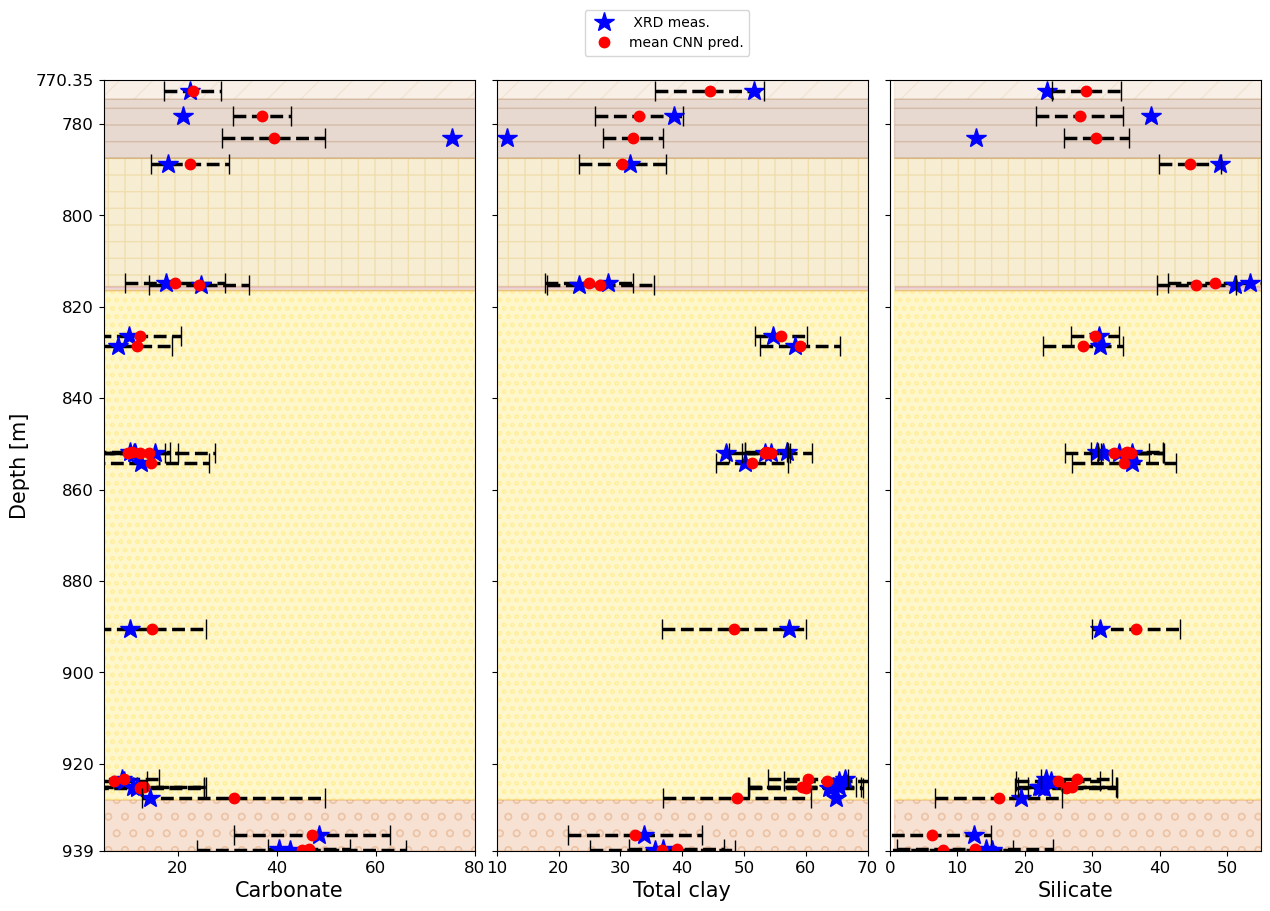}
    \caption{Mean (red dots) and 2$\sigma$ of the CNN predicted mineral content from images at XRD depth $\pm$ 5cm and the according bulk XRD measurements (blue stars).} 
    \label{fig:meas_model}
\end{figure}

In total 23 mineralogical data points obtained by XRD measurements on samples from core intervals used in image analysis were available for comparison with the model predictions. 
The comparison of core and log mineral compositions needs to consider small differences and uncertainties. For example, the rock samples taken for the analysis represent a core fragment with a volume of 1-10 cm\textsuperscript{3}. These samples are not necessarily located on the surface of the core. Quite in contrary, the samples are taken from the core center to avoid contamination or any alteration processes. Thus, the prediction based on the visual information from the surface can not be more accurate than the typical variation of mineralogical content within a domain of  10$\times$10$\times$10 cm\textsuperscript{3}, which was taken for the laboratory analysis.  Accordingly, 11 image segments were taken for each data point as input for the model and the mean and standard deviation, $\sigma$, of the predicted mineral content were computed. The full table of results can be found in the appendix \ref{App:XRD}, Table \ref{tab:meas_model}. In Figure \ref{fig:meas_model} the mean (red dot) and the 95\% confidence interval, i.e. the 2$\sigma$ region, of predictions are depicted for each mineral, together with the real measurements (blue stars). Almost all the measurements lie within the 2$\sigma$ region. The only significant deviation is observed in one sample from Humphriesioolith. The photographic interpretation shows an unusually high carbonate content and is part of a thin layer of elevated carbonate content present within the heterogeneous Humphriesioolith unit.

To have a comparison to the MultiMin log data, also for the predicted model data the Spearman correlation coefficient ($cc$) was calculated for each mineral, taking the prediction at the MultiMin log data depth: 

\noindent Carbonate: $cc_{Multimin}=0.80$ ($p_{val}=4.21E-06$), $cc_{CNN}=0.90$ ($p_{val}=6.90E-09$), 

\noindent Clay: $cc_{Multimin}=0.93$ ($p_{val}=1.42E-10$), $cc_{CNN}=0.90$ ($p_{val}=6.43e-09$), 

\noindent Silicate: $cc_{Multimin}=0.84$ ($p_{val}=5.51E-07$), $cc_{CNN}=0.81$ ($p_{val}=3.31E-06$). 

\begin{figure}
    \centering
    \includegraphics[width = \textwidth ]{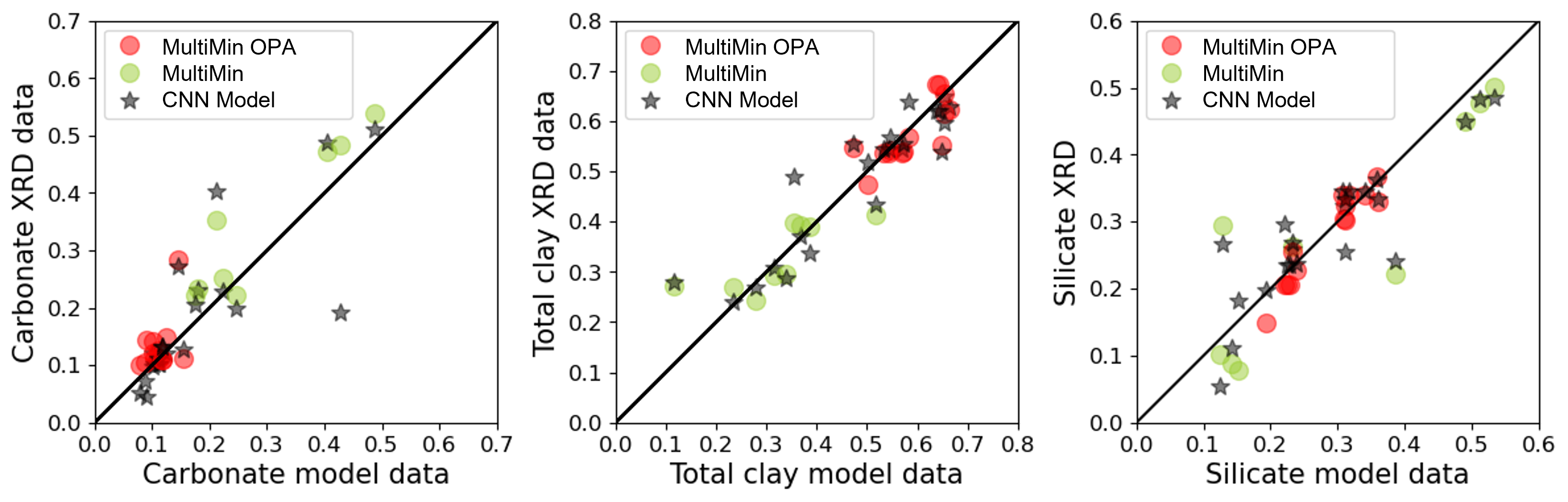}
    \caption{Bulk XRD measurements (only representing carbonate, total clay, and silicate compositions) plotted versus MultiMin log data (red and green dots) and overlain with CNN predicted mineral content at MultiMin log data depth (black stars). The red dots correspond to the Opalinus clay, whereas the green dots to all the other formations.}
    \label{fig:meas_pred_CNN_Multimin_OPA}
\end{figure}

\noindent All values, including the relative and absolute errors; can be found in the appendix \ref{App:XRD}, \ref{tab:meas_model_metric}.  
This metric was chosen, since this was the metric used in \cite{NAB20_09_X} to compare the MultiMin log data to the XRD measurements. The neural network model predictions are in good agreement with the MultiMin log data and show similar accuracy as illustrated in Figure \ref{fig:meas_pred_CNN_Multimin_OPA}. 

In summary, the mineral content regression model presents a promising tool for mineral estimation from drill core images for the selected formations within this core, although further refinement of the model to increase the accuracy would be desired. In the future, an expansion of the dataset and inclusion of different lithological core sections will be considered. 
\section{Conclusion and Outlook}\label{conclusion}
This study successfully established and tested an automatic workflow for preprocessing and analysis of drill core images based on machine learning methods. A pretrained ResNet architecture was trained as part of the workflow to classify the 1cm drill core segments into 6 different formation classes (for which data and images were available). The classification model achieved a prediction accuracy of 96.7\% for unseen images. Big parts of this formation classification model were subsequently used as a backbone architecture to establish a NN model to predict the mineral content (silicate, total clay and carbonate) from only drill core images. The CNN mineral content regression predictions were compared to the XRD measurements and showed comparable good correlations as the MultiMin log data. The critical technique used for both models was transfer learning, which involved the usage of models that were trained on related tasks and hence, the re-use of the obtained knowledge. 
It has to be emphasized that due to the limited number of experimentally measured mineralogical data used in this study (23), the CNN model was trained using a model based on the MultiMin log data set (361) derived from drill log data. The data analysis shows that both models, CNN and MultiMin log, demonstrate the same prediction performance when benchmarked against the measured mineralogical lab (XRD) data.  In contrast to other models in the literature, the CNN model constructed in this study relies on images only and can be used for the image-based interpolation of mineralogical data down to 1 cm resolution. 

Although both presented CNN models (formation classification, mineral content regression) show already good accuracy, there are several options to possibly further improve the performance, which lie outside the scope of the current paper:
First, the hyperparameters, so the number of layers and neurons, as well as the activation functions were chosen to be fixed. A thorough search for optimal hyperparameters could drastically increase the performance of the model. 
Second, another choice of the backbone architecture (here ResNet was used) could improve the model and is planned for future work (like e.g. transformer networks or self-attention mechanisms).

In general, the transfer learning strategy from pretrained models seems to perform better than starting training from scratch. The ideal case would be to have a model that was trained on a huge data set of drill core images from different places all over the world to extract already important features from the images. The model fine-tuning could then be done for the specific applications and data sets. 

The number of data points is essential for the performance of a machine learning model. In this study, the formation classification model is based on more than 5000 images, whereas the mineral content regression model was based on only 361 images. Increasing the amount of data would definitely improve the model and make it more robust and capable of generalization. 

The data set in this study could benefit from further augmentation by incorporating images from other boreholes of the same formations of complementary geological units. This expansion would not only increase the data set's size but also enhance the model's ability to generalize across diverse and possibly challenging geological formations and thus rock compositions. Another cost-efficient option to augment the data set would be to create a synthetic data set with geostatistical methods. 

The performance of the models might also be influenced by the choice of the segmentation size of the images, in this work we used 1 cm image segments. Decreasing the width or height of the segments would be a way of enlarging the data set, although each segment would have less information. Increasing the width would have the opposite effect, the data set available would be reduced, but each segment would contain more information. Optimizing this aspect can be crucial for achieving a balance between data quantity and quality. 

To enhance the data quality, the radial distortion of the images, due to the position of the camera in relation to the drill core, should be considered. In future work, the implementation of a transformation function for correcting the radial distortion is planned. 

In summary, this study provides a solid foundation for the regression of mineral content from drill core images using deep learning techniques. The outlined future research directions provide a road map for improving model performance, data set robustness, and overall predictive accuracy. Ultimately image-based regression methodologies will have the potential to support the field of geological analysis and drilling technologies.

\section*{Declarations}

\subsection{Ethics approval and consent to participate}
Not applicable

\subsection{Consent for publication}
All authors have given their consent for publication.

\subsection{Availability of data and material}
The data that support the findings of this study are available from Nagra but restrictions apply to the availability of these data, which were used under license for the current study, and so are not publicly available. Data are however available from the authors upon reasonable request and with permission of Nagra. The code framework is available from the authors upon reasonable request. 

\subsection{Competing interests}
The authors declare no competing interests. 

\subsection{Funding}
EURAD-DONUT WP received funding from the European Union’s Horizon
2020 research and innovation programme under grant agreement No. 847593.

\subsection{Authors' contributions}
RB, SC and NP conceptualized the work. RB, ILB and NP designed the methodology. RB and ILB developed the software framework and validated the models. RW provided the data and accompanying information. GK contributed to the data analysis and the data and model interpretation. RB drafted the first version of the manuscript. SC, GK, ILB, RW and NP substantially contributed to the manuscript. All authors read and approved the final manuscript.

\subsection{Acknowledgements}
We sincerely thank Marc Pollefeys (ETHZ) for his insights and discussions regarding the implementation of neural networks in computer vision. We appreciate the internship of Geeta Goyal, that gave a starting point for this work. We are thankful for thoughtful discussions with Maximilian Mandl (Nagra) beforehand on the usage of machine learning for drill cores. Thanks also go the the Database Managment Group at Nagra which provided maps and core images. We appreciate a lot the enlightening conversations with Thomas Gimmi (PSI) on the available datasets.

\newpage

\begin{appendices}

\section{}
\subsection{Appendix List of Abbreviations (alphabetical order)} \label{App:Abbreviations}
\begin{itemize}
\item cc: correlation coefficient
\item chemical elements: U: Uranium; Th: Thorium; K: Potassium; Fe: Iron; Si: Silicon; Ca: Calcium; Al: Aluminium; Ti: Titanium; S: Sulfur; 
\item CNN: Convolutional neural network
\item ICC: International Color Consortium (colour profile) 
\item NN: Neural network
\item $p_{val}$: P-value (probability)
\item pXRF: Portable X-ray Fluorescence analysis
\item R2: Coefficient of determination
\item ReLU: Rectified Linear Unit
\item ResNet: Residual Neural Network
\item RF: Random Forest
\item SVM: Support Vector Machine
\item TIFF: Tagged Image File Format
\item VGG: Visual Geometry Group (a deep CNN architecture)
\item XRD: X-ray Diffraction
\end{itemize}
\newpage
\pagebreak

\subsection{Appendix Additional figures for mineral content regression}\label{App:regression}
\begin{figure}[ht]
    \centering
    \includegraphics[width=\textwidth]{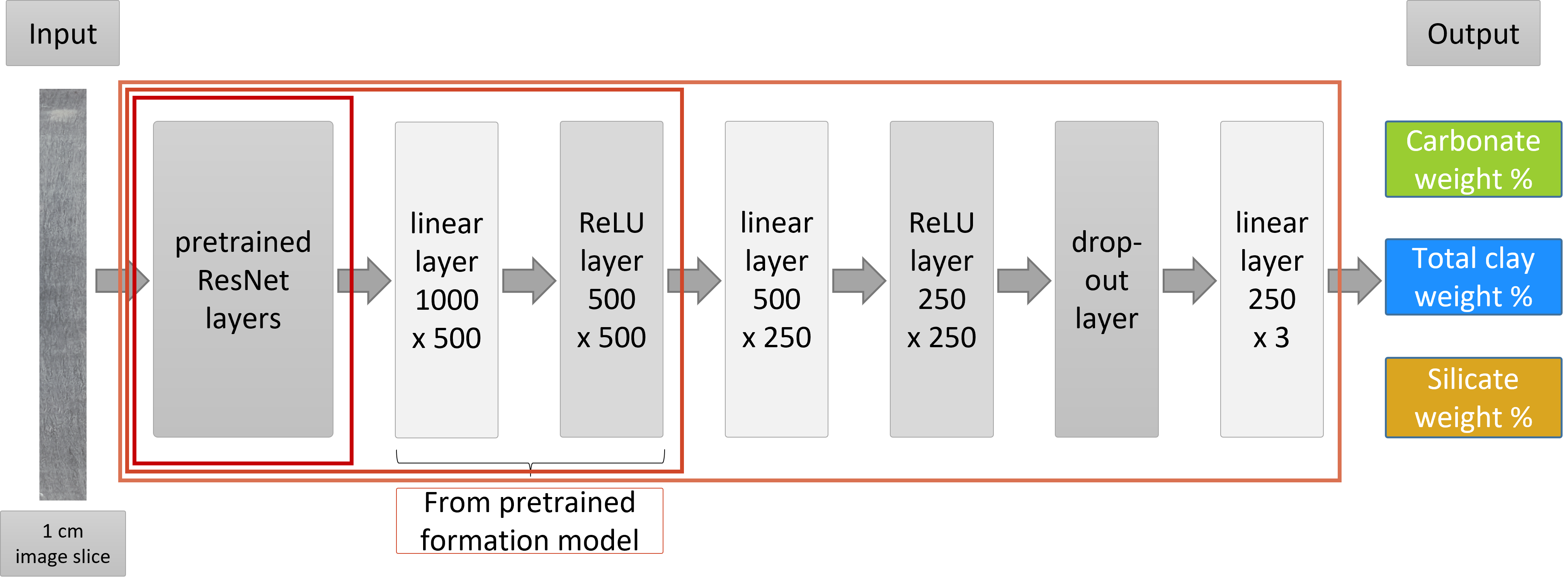}
    \caption{Architecture of the neural network for the mineral content regression with one formation model as a backbone architecture.}
    \label{fig:CNN-architecture}
\end{figure}

\begin{figure}[ht]
    \centering    \includegraphics[width = 0.7\textwidth]{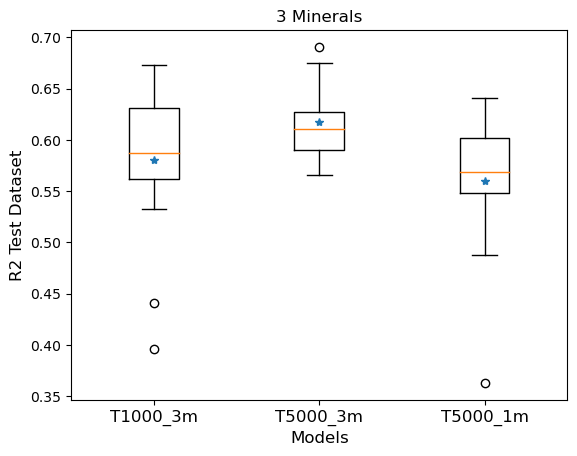}
    \caption{Boxplot for comparing the three different model types for the mineral content regression in terms of the R2 values of the test data set.} \label{fig:boxplot_regression_appendix}
\end{figure}

\begin{figure}[ht]
    \centering
    \includegraphics[width=\textwidth]{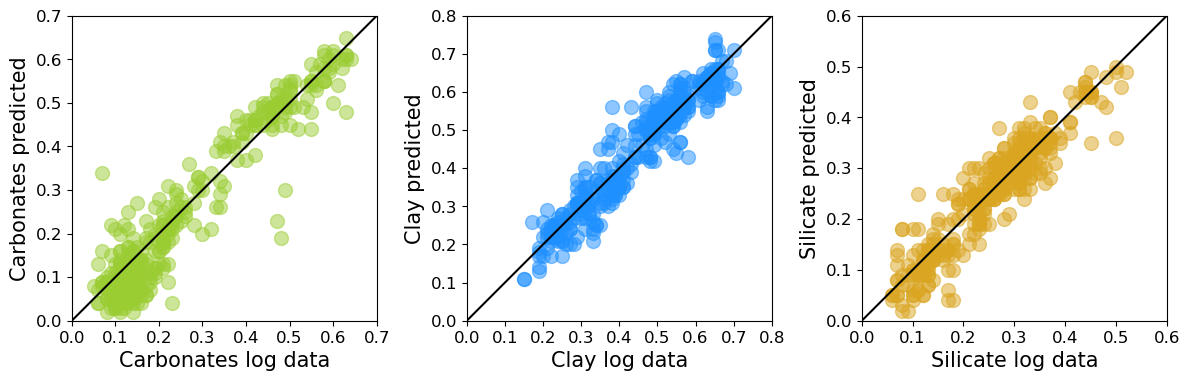}
    \caption{MultiMin log data vs CNN model predicted data for each mineral separately, the x and y axis show the according weight \%. }
    \label{fig:meas_pred_single_mineral}
\end{figure}

\pagebreak

\clearpage

\newpage

\subsection{Appendix: Additional tables for comparison between model and bulk XRD measurements}\label{App:XRD}
\begin{table}[ht]
\begin{tabular}{|l|l|l|l|l|l|l|l|l|l|l|l|}
\hline
            & \textbf{\begin{tabular}[c]{@{}r@{}}Mean \\ depth \\ {[}m{]}\end{tabular}} & \textbf{Sil.} & \textbf{\begin{tabular}[c]{@{}r@{}}Sil.\\ pred \\ mean\end{tabular}} & \textbf{\begin{tabular}[c]{@{}r@{}}Sil.\\ pred\\  std\end{tabular}} & \textbf{Carb.} & \textbf{\begin{tabular}[c]{@{}r@{}}Carb.\\ pred \\ mean\end{tabular}} & \textbf{\begin{tabular}[c]{@{}r@{}}Carb.\\ pred\\  std\end{tabular}} & \textbf{Clay} & \textbf{\begin{tabular}[c]{@{}r@{}}Clay\\ pred\\  mean\end{tabular}} & \textbf{\begin{tabular}[c]{@{}r@{}}Clay\\ pred\\  std\end{tabular}} & \textbf{\begin{tabular}[c]{@{}r} Lith-\\ology \end{tabular}} \\ \hline
\textbf{0}  & 772.73                      & 23.3          & 29.16                   & 2.58                   & 22.4           & 23.01                    & 2.88                    & 51.66         & 44.43                   & 4.41                   & Park. Würt. Sch.   \\ \hline
\textbf{1}  & 778.32                      & 38.7          & 28.14                   & 3.25                   & 21.1           & 36.99                    & 2.93                    & 38.71         & 32.98                   & 3.54                   & Humphr. Form.      \\ \hline
\textbf{2}  & 783.13                      & 12.8          & 30.63                   & 2.38                   & 75.34          & 39.36                    & 5.2                     & 11.62         & 32.02                   & 2.44                   & Humphr. Form.      \\ \hline
\textbf{3}  & 788.76                      & 49            & 44.52                   & 2.3                    & 17.98          & 22.41                    & 3.95                    & 31.6          & 30.33                   & 3.54                   & Wedels. Form.      \\ \hline
\textbf{4}  & 814.82                      & 53.4          & 48.2                    & 3.47                   & 17.52          & 19.4                     & 5.05                    & 27.93         & 24.97                   & 3.56                   & Wedels. Form.      \\ \hline
\textbf{5}  & 815.28                      & 51.2          & 45.46                   & 2.93                   & 24.68          & 24.33                    & 5.04                    & 23.32         & 26.74                   & 4.32                   & Wedels. Form.      \\ \hline
\textbf{6}  & 826.39                      & 31            & 30.44                   & 1.77                   & 10.17          & 12.28                    & 4.19                    & 54.68         & 56                      & 2.09                   & Opalinus Clay      \\ \hline
\textbf{7}  & 828.51                      & 31.17         & 28.67                   & 2.97                   & 7.87           & 11.68                    & 3.53                    & 58.24         & 59.01                   & 3.22                   & Opalinus Clay      \\ \hline
\textbf{8}  & 851.84                      & 30.66         & 35.15                   & 2.7                    & 10.41          & 10.66                    & 3.87                    & 57            & 53.53                   & 1.69                   & Opalinus Clay      \\ \hline
\textbf{9}  & 851.88                      & 31.65         & 35.73                   & 2.4                    & 11.27          & 9.87                     & 3.75                    & 54.37         & 53.77                   & 1.82                   & Opalinus Clay      \\ \hline
\textbf{10} & 851.95                      & 33.99         & 34.93                   & 1.77                   & 11.04          & 14.19                    & 6.62                    & 53.4          & 53.36                   & 1.87                   & Opalinus Clay      \\ \hline
\textbf{11} & 852.05                      & 35.97         & 33.3                    & 3.66                   & 15.43          & 12.26                    & 3.92                    & 47.16         & 54.28                   & 3.39                   & Opalinus Clay      \\ \hline
\textbf{12} & 854.18                      & 35.89         & 34.72                   & 3.84                   & 12.49          & 14.49                    & 5.94                    & 50.11         & 51.23                   & 2.92                   & Opalinus Clay      \\ \hline
\textbf{13} & 890.39                      & 31.2          & 36.48                   & 3.26                   & 10.25          & 14.73                    & 5.46                    & 57.31         & 48.36                   & 5.84                   & Opalinus Clay      \\ \hline
\textbf{14} & 923.34                      & 23.2          & 27.72                   & 2.64                   & 8.67           & 9.2                      & 3.5                     & 66.37         & 60.36                   & 3.25                   & Opalinus Clay      \\ \hline
\textbf{15} & 923.81                      & 23.89         & 24.99                   & 3.11                   & 9.03           & 7.19                     & 3.33                    & 65.41         & 63.46                   & 3.53                   & Opalinus Clay      \\ \hline
\textbf{16} & 925.11                      & 22.11         & 27.02                   & 3.28                   & 10.96          & 13.19                    & 6.28                    & 65.41         & 59.32                   & 4.37                   & Opalinus Clay      \\ \hline
\textbf{17} & 925.16                      & 22.39         & 26.36                   & 3.69                   & 11.7           & 12.46                    & 6.43                    & 64.35         & 59.84                   & 4.55                   & Opalinus Clay      \\ \hline
\textbf{18} & 925.17                      & 22.86         & 26.12                   & 3.79                   & 11.73          & 12.41                    & 6.43                    & 63.71         & 59.99                   & 4.6                    & Opalinus Clay      \\ \hline
\textbf{19} & 927.41                      & 19.38         & 16.12                   & 4.71                   & 14.4           & 31.27                    & 9.23                    & 64.86         & 48.86                   & 5.96                   & Opalinus Clay      \\ \hline
\textbf{20} & 935.59                      & 12.52         & 6.24                    & 4.37                   & 48.63          & 47.13                    & 7.84                    & 33.77         & 32.39                   & 5.43                   & Staffel. Form.     \\ \hline
\textbf{21} & 938.53                      & 14.2          & 12.6                    & 5.8                    & 40.43          & 46.45                    & 4.15                    & 36.85         & 39.13                   & 3.85                   & Staffel. Form.     \\ \hline
\textbf{22} & 938.9                       & 15.2          & 7.93                    & 5.14                   & 42.77          & 45.05                    & 10.55                   & 35.6          & 36.77                   & 5.85                   & Staffel. Form.     \\ \hline
\end{tabular}

\caption{Comparison of model predicted data and bulk XRD measured data. The mean depth is given in meter [m]. The columns Sil., Carb., Clay show the measured mineral content. The columns Sil. pred.mean, Carb.pred.mean, Clay pred.mean and Sil.pred.std., Carb.pred.std., Clay pred.std. denote the mean and standard deviation of the predictions with the CNN model for images of the according measurement depth $\pm 5$cm.  }\label{tab:meas_model}
\end{table}

\begin{table}[]
\begin{tabular}{|l|r|r|r|r|r|r|}
\hline & \textbf{\begin{tabular}[c]{@{}r@{}}XRD\\ MultiMin\\ Carbonate\end{tabular}} & \textbf{\begin{tabular}[c]{@{}r@{}}XRD\\ MultiMin\\ Total clay\end{tabular}} & \textbf{\begin{tabular}[c]{@{}r@{}}XRD\\ MultiMin\\ Silicate\end{tabular}} & \textbf{\begin{tabular}[c]{@{}r@{}}XRD\\ CNN\\ Carbonate\end{tabular}} & \textbf{\begin{tabular}[c]{@{}r@{}}XRD\\ CNN\\ Total clay\end{tabular}} & \textbf{\begin{tabular}[c]{@{}r@{}}XRD\\ CNN\\ Silicate\end{tabular}} \\ \hline
Abs.err. & 5.36  & 3.91 & 3.81  & 5.69   & 4.26 & 3.87 \\ \hline
Rel.err. & 25.29\% & 12.84\%  & 17.88\% & 24.08\% & 13.59\% & 17.38\%\\ \hline
R2 & 0.68  & 0.88  & 0.75  & 0.61  & 0.84  & 0.78  \\ \hline
cc & 0.80  & 0.93  & 0.84  & 0.90  & 0.90 & 0.81 \\ \hline
p-val & 4.21E-06 & 1.42E-10 & 5.51E-07 & 6.90E-09 & 6.43E-09 & 3.31E-06 \\ \hline
\end{tabular}\label{tab:meas_model_metric}
\caption{Absolute (Abs.err.), relative (Rel.err.) error, coefficient of determination (R2) and Spearman correlation coefficient (cc) of the three minerals and the according p-values (p-val), computed for the 23 XRD measurement data points compared to the MultiMin Log data and the CNN model predictions. The metrics are computed for the predicted values at the corresponding MultiMin Log data depth. }
\end{table}

\newpage
\pagebreak

\subsection{Appendix Overview of core photographs}\label{App:Corephoto}
The following section contains the core images available and used in this study, already preprocessed until step 3, so the original image was color and white balance corrected and cropped from the background. For each photo, the specific depth range is written above the photo together with the corresponding lithology class. The blue rectangles indicate where XRD laboratory measurements were performed. For model validation purposes, the slices at the depth given for the XRD measurements $\pm$ 5cm are considered (blue rectangles).
\newpage


\section*{Supplementary material}

**Drill core photographs, extracted from the background**

770.35m – 770.92m, Parkinsoni-Württembergica Schichten

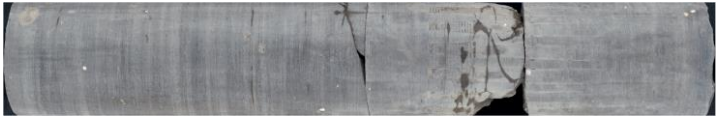

770.92m-771.75m, Parkinsoni-Württembergica Schichten

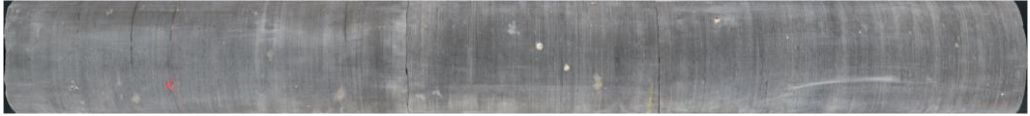

771.75m-772.58m, Parkinsoni-Württembergica Schichten

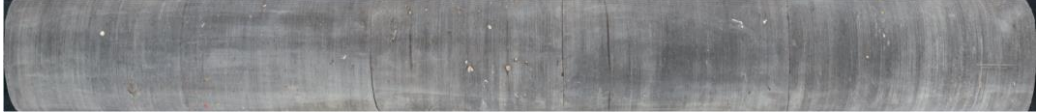

772.58m-773.06m, Parkinsoni-Württembergica Schichten

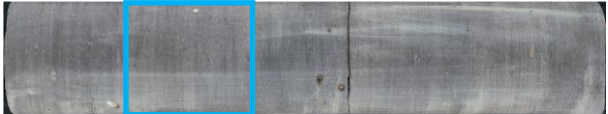

773.06m-773.55m, Parkinsoni-Württembergica Schichten

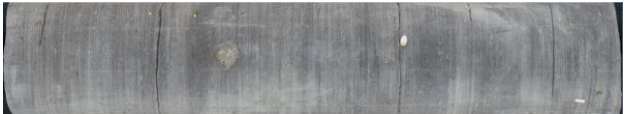

773.55m-773.73m, Parkinsoni-Württembergica Schichten

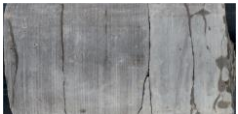

773.73m-773.80m, Parkinsoni-Württembergica Schichten

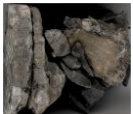

773.80m-773.93m, Parkinsoni-Württembergica Schichten

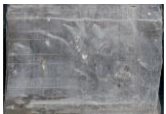

773.93m-774.39m, Parkinsoni-Württembergica Schichten

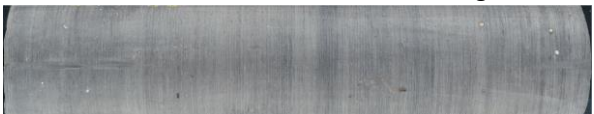

774.39m-774.55m, Parkinsoni-Württembergica Schichten 774.55m-774.83m Humphriesoolith Formation

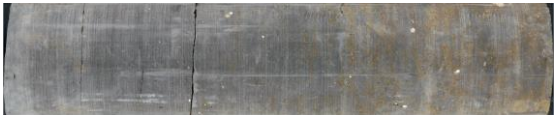

774.83m-775.23m, Humphriesoolith Formation

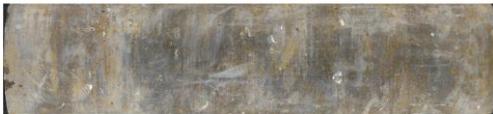

775.23m-775.63m, Humphriesoolith Formation

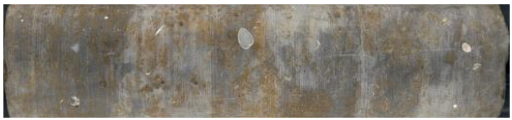

**Drill core photographs, extracted from the background**

775.63m-776.36m, Humphriesoolith Formation

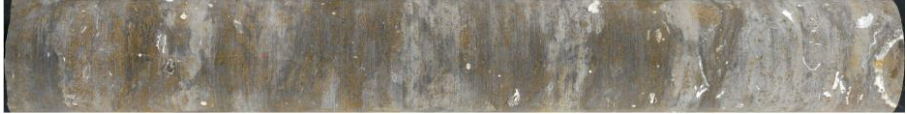

776.36m-776.92m, Humphriesoolith Formation

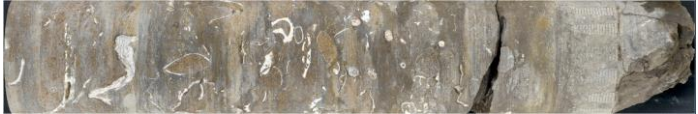

776.92m-777.80m, Humphriesoolith Formation

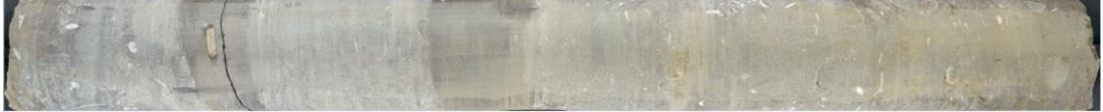

777.80m-778.22m, Humphriesoolith Formation

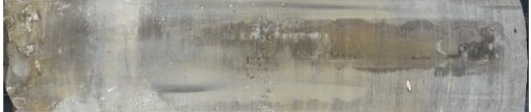

778.22m-778.54m, Humphriesoolith Formation

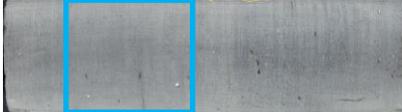

778.54m-779.44m, Humphriesoolith Formation

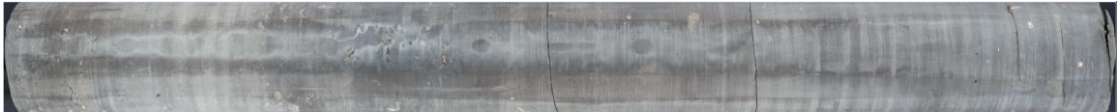

779.44m-779.93m, Humphriesoolith Formation

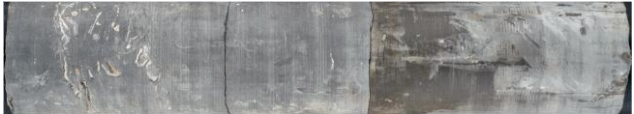

779.44m-779.93m, Humphriesoolith Formation

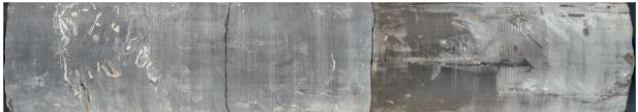

779.93m-780.63m, Humphriesoolith Formation

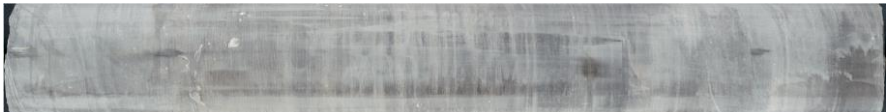

782.94m-783.76m, Humphriesoolith Formation

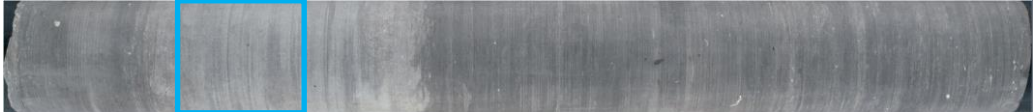

783.76m-784.60m, Humphriesoolith Formation

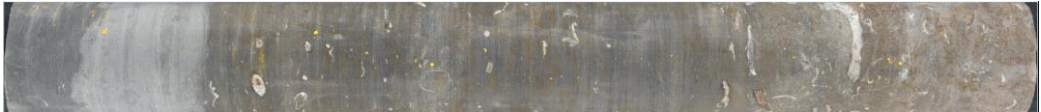

784.60m-785.20m, Humphriesoolith Formation

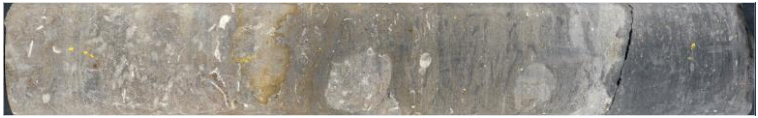

**Drill core photographs, extracted from the background**

785.20m-785.94m, Humphriesoolith Formation
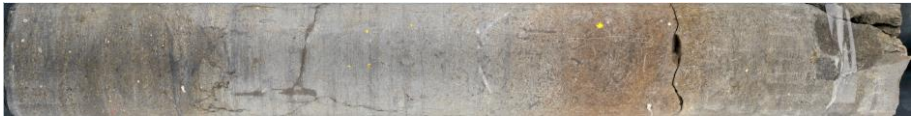

785.94m-786.74m, Humphriesoolith Formation
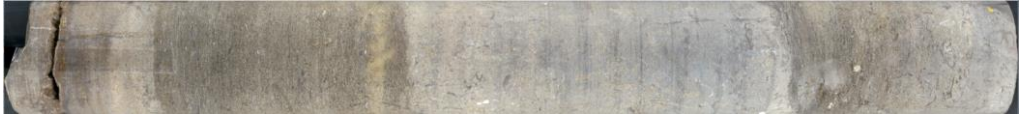

786.74m-787.50m, Humphriesoolith Formation; 787.50m-787.63m Wedelsandstein Formation
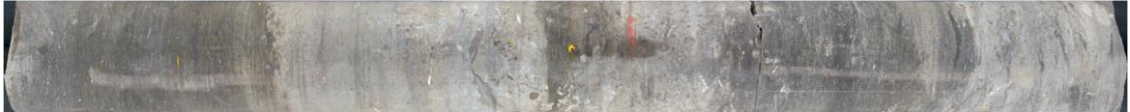

787.63m-788.43m, Wedelsandstein Formation
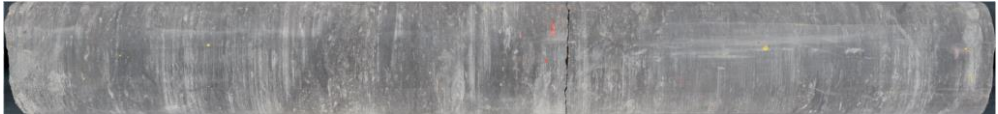

788.43m-788.99m, Wedelsandstein Formation
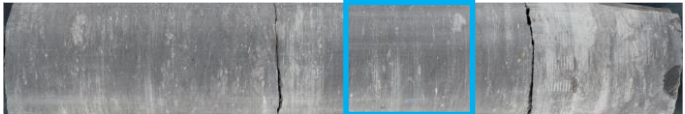

788.99m-789.92m, Wedelsandstein Formation
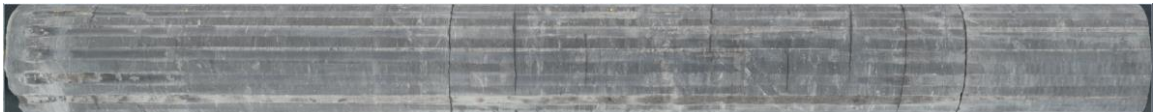

814.70m-815.51m, Wedelsandstein Formation; 815.51m-815.38m, Murchisonae-Oolith Formation
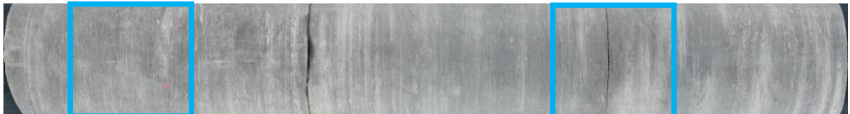

815.38m-815.68m, Murchisonae-Oolith Formation
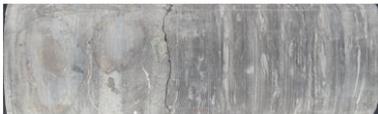

815.68m-816.02m, Murchisonae-Oolith Formation
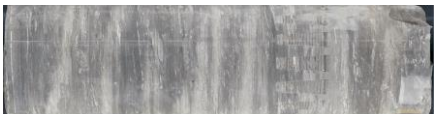

816.02m-816.42m, Murchisonae-Oolith Formation
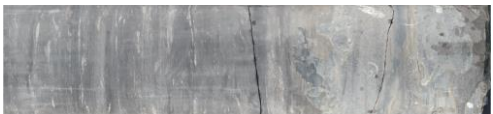

816.42m-817.32m, Opalinus Clay
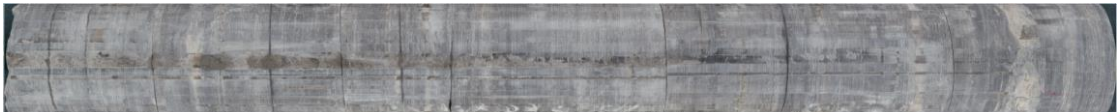

817.32m-818.04m, Opalinus Clay
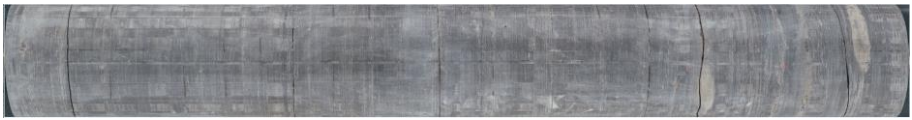

**Drill core photographs, extracted from the background**

818.04m-818.63m, Opalinus Clay

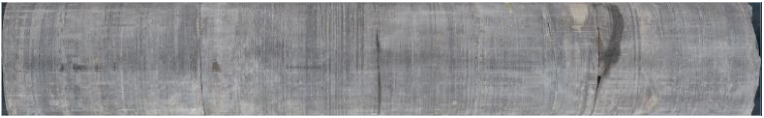

818.63m-819.38m, Opalinus Clay

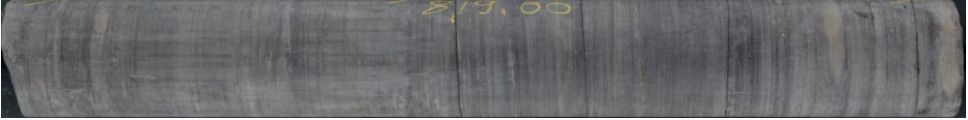

819.38m-819.41m, Opalinus Clay

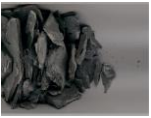

819.41m-819.56m, Opalinus Clay

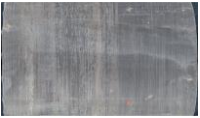

819.56m-819.75m, Opalinus Clay

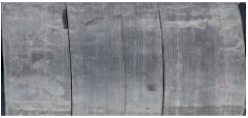

819.75m-820.15m, Opalinus Clay

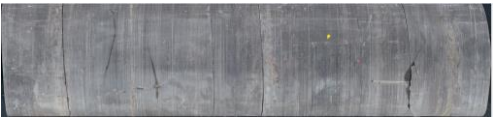

824.04m-824.42m, Opalinus Clay

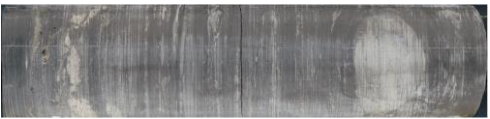

824.42m-825.40m, Opalinus Clay

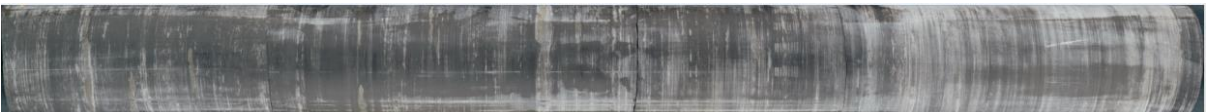

825.40m-825.57m, Opalinus Clay

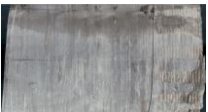

825.57m-825.66m, Opalinus Clay

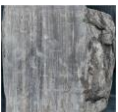

825.66m-826.48m, Opalinus Clay

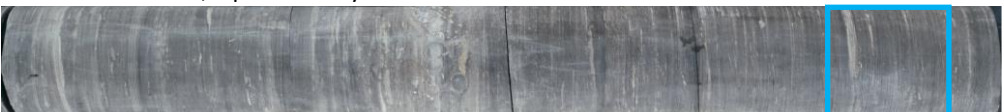

826.48m-826.82m, Opalinus Clay

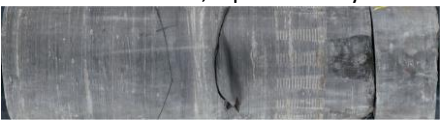

**Drill core photographs, extracted from the background**

826.82m-827.02m, Opalinus Clay

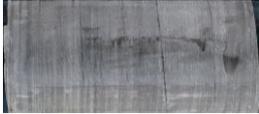

827.02m-827.28m, Opalinus Clay

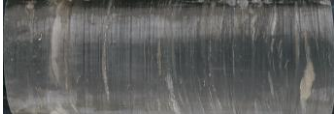

827.28m-827.75m, Opalinus Clay

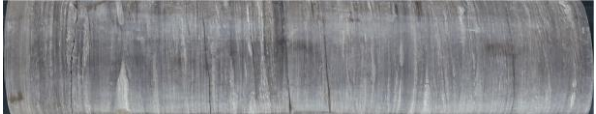

827.75m-828.38m, Opalinus Clay

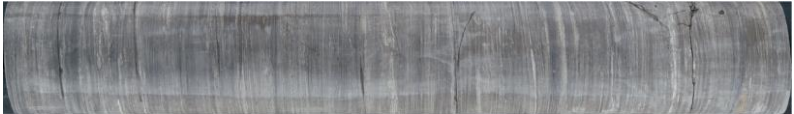

828.38m-828.70m, Opalinus Clay

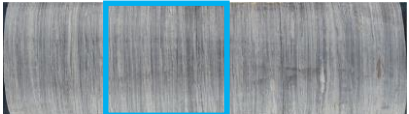

850.04m-850.95m, Opalinus Clay

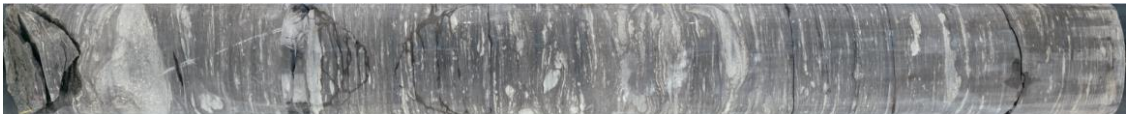

850.95m-851.70m, Opalinus Clay

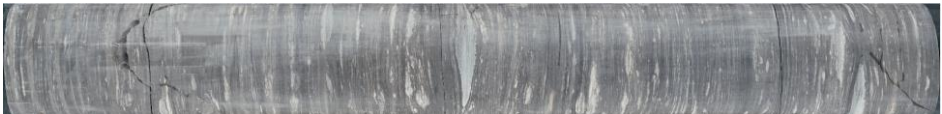

851.70m-852.20m, Opalinus Clay

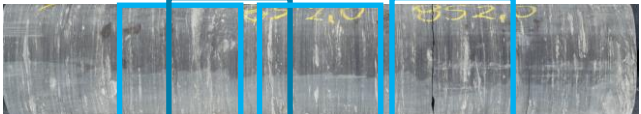

852.20m-852.30m, Opalinus Clay

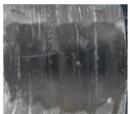

852.30m-853.13m, Opalinus Clay

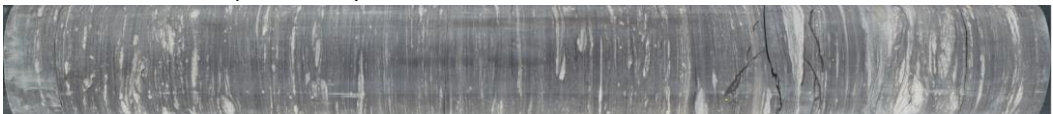

853.13m-853.36m, Opalinus Clay

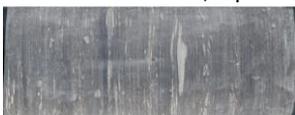

853.36m-853.60m, Opalinus Clay

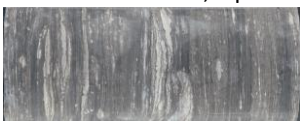

**Drill core photographs, extracted from the background**

853.60m-853.75m, Opalinus Clay

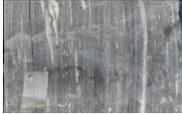

853.75m-853.90m, Opalinus Clay

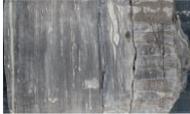

853.90m-854.05m, Opalinus Clay

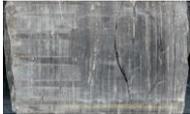

854.05m-854.30m, Opalinus Clay

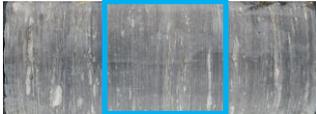

854.30m-854.84m, Opalinus Clay

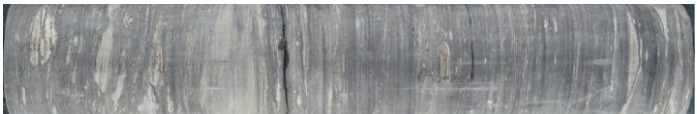

854.84m-855.47m, Opalinus Clay

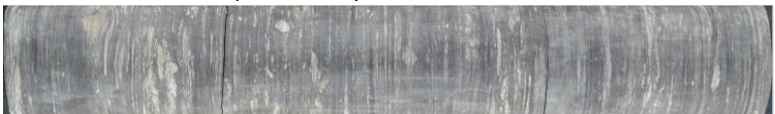

860.03m-860.85m, Opalinus Clay

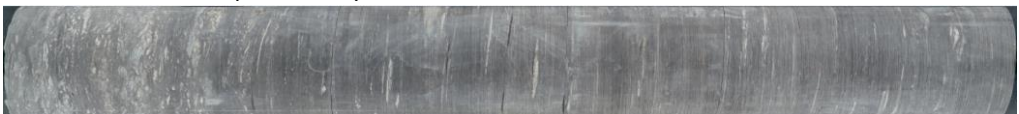

860.85m-861.22m, Opalinus Clay

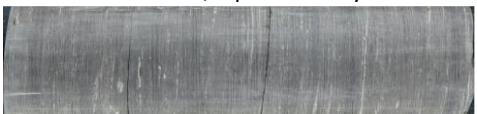

861.22m-861.54m, Opalinus Clay

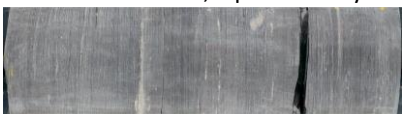

861.54m-862.08m, Opalinus Clay

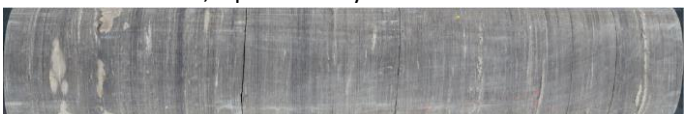

862.08m-862.39m, Opalinus Clay

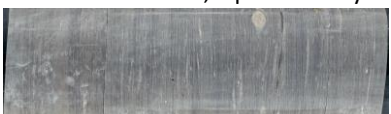

862.39m-862.88m, Opalinus Clay

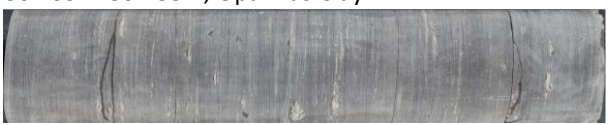

**Drill core photographs, extracted from the background**

862.88m-862.95m, Opalinus Clay

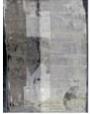

862.95m-863.56m, Opalinus Clay

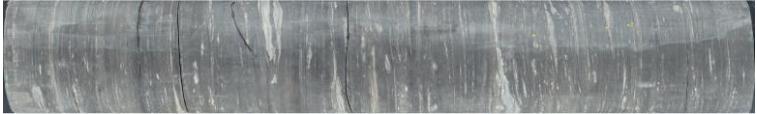

863.56m-864.00m, Opalinus Clay

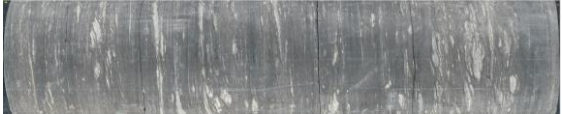

864.00m-864.62m, Opalinus Clay

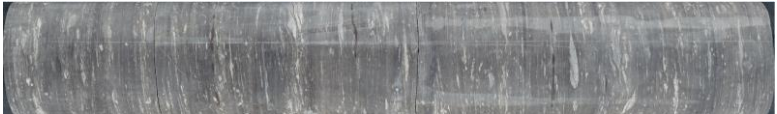

887.15m-887.83m, Opalinus Clay

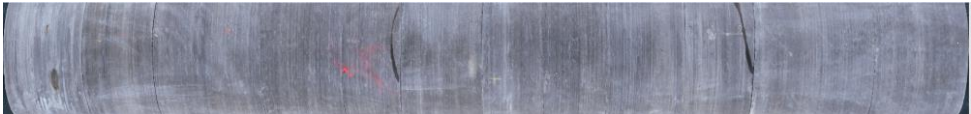

887.73m-888.13m, Opalinus Clay

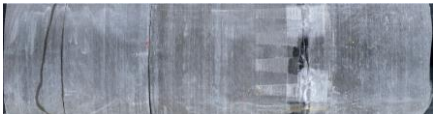

888.13m-889.09m, Opalinus Clay

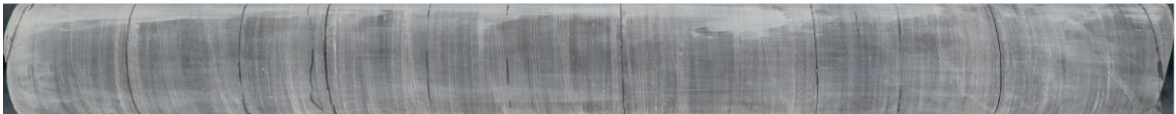

889.09m-889.97m, Opalinus Clay

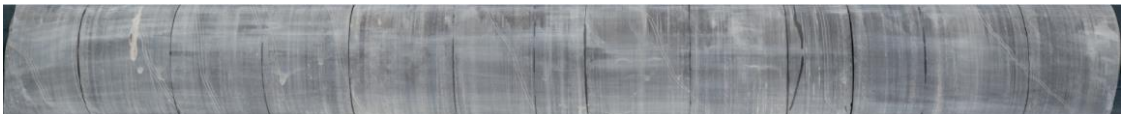

889.97m-890.50m, Opalinus Clay

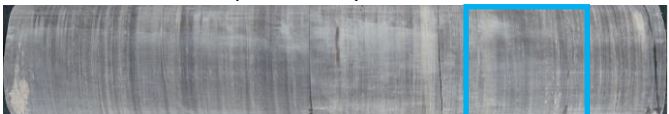

890.50m-890.85m, Opalinus Clay

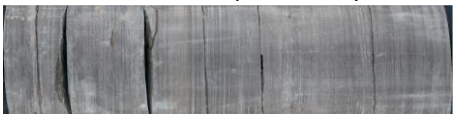

890.85m-891.03m, Opalinus Clay

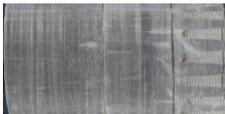

891.03m-891.89m, Opalinus Clay

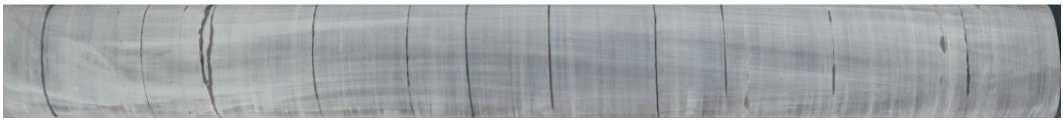

**Drill core photographs, extracted from the background**

922.49m-923.24m, Opalinus Clay

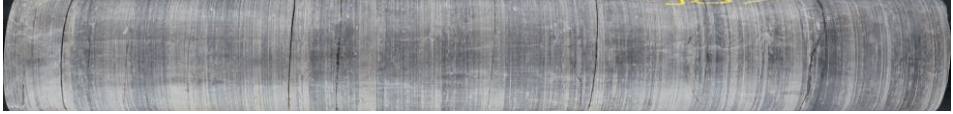

923.24m-923.94m, Opalinus Clay

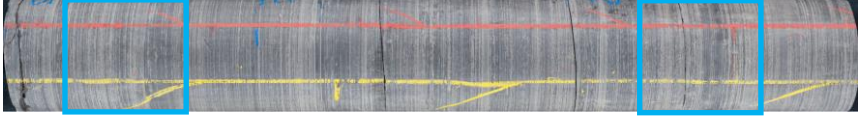

923.94m-924.08m, Opalinus Clay

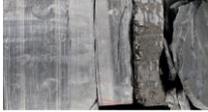

924.08m-924.96m, Opalinus Clay

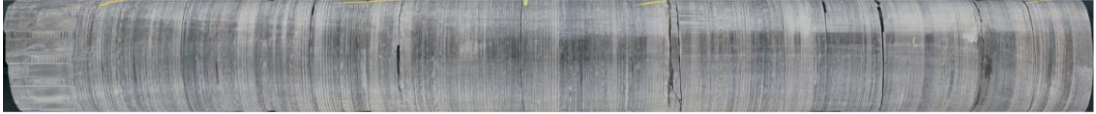

924.96m-925.41m, Opalinus Clay

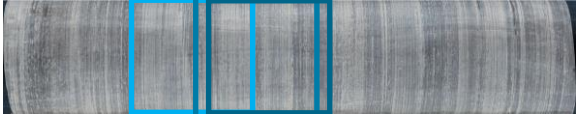

925.41m-926.03m, Opalinus Clay

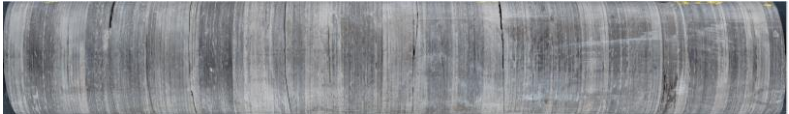

926.03m-926.70m, Opalinus Clay

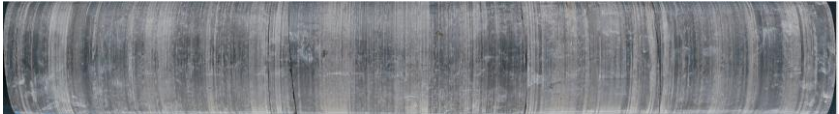

926.70m-927.15m, Opalinus Clay

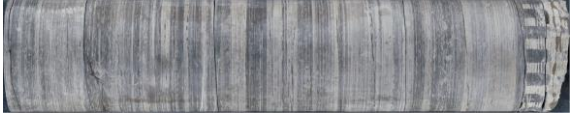

927.15m-927.53m, Opalinus Clay

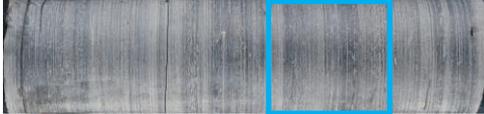

927.53m-927.91m Opalinus Clay; 927.91m-928.31m Staffelegg Formation

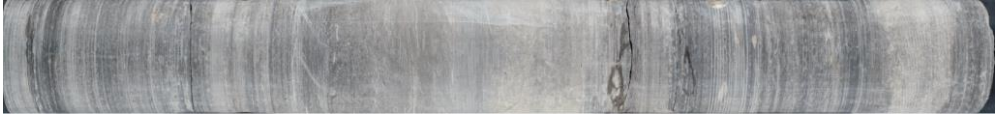

932.00m-932.96m, Staffelegg Formation

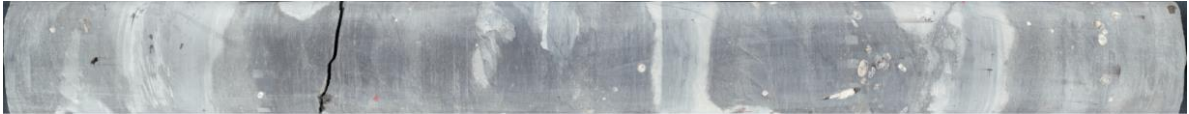

932.96m-933.20m, Staffelegg Formation

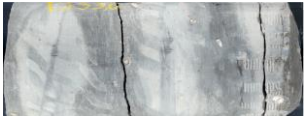

**Drill core photographs, extracted from the background**

933.20m-933.29m, Staffelegg Formation

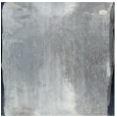

933.29m-934.16m, Staffelegg Formation

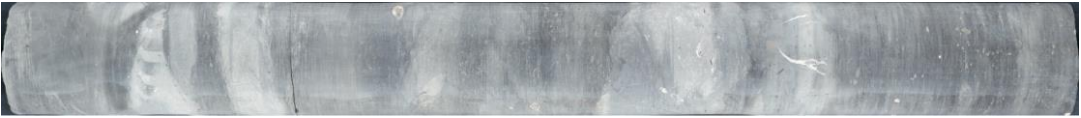

934.16m-935.03m, Staffelegg Formation

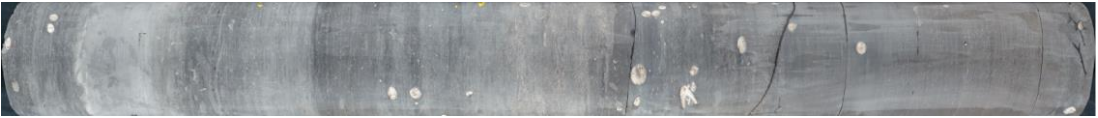

935.03m-935.31m, Staffelegg Formation

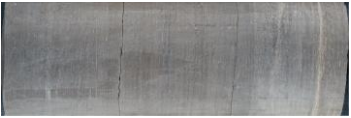

935.31m-935.94m, Staffelegg Formation

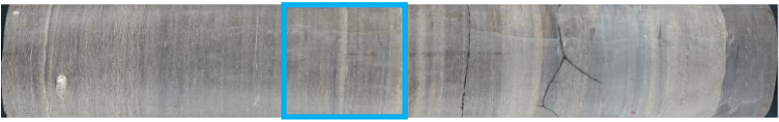

935.94m-936.18m, Staffelegg Formation

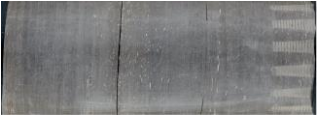

936.18m-937.17m, Staffelegg Formation

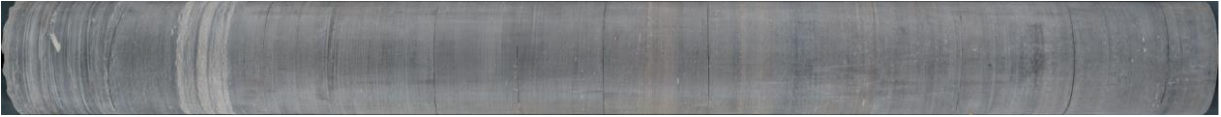

937.17m-938.13m, Staffelegg Formation

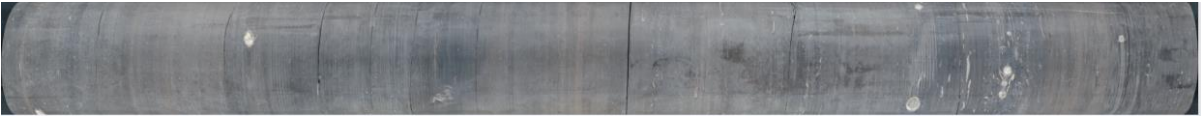

938.13m-939.01m, Staffelegg Formation

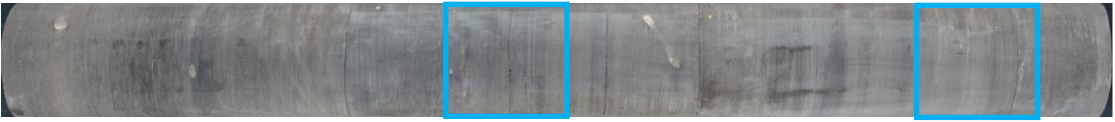


\end{appendices}


\newpage
\pagebreak

\bibliography{refs}% common bib file

\end{document}